\documentclass{article} %
\usepackage[table]{xcolor}   %
\usepackage{iclr2027_conference,times}
\usepackage[T1]{fontenc}   %

\usepackage{amsmath,amsfonts,bm}
\usepackage{dashrule}

\def\eqref#1{equation~\ref{#1}}

\def\1{\bm{1}}

\DeclareMathAlphabet{\mathsfit}{\encodingdefault}{\sfdefault}{m}{sl}
\SetMathAlphabet{\mathsfit}{bold}{\encodingdefault}{\sfdefault}{bx}{n}

\newcommand{\R}{\mathbb{R}}

\newcommand{\method}{{\sc rOUT}\xspace}

\newcommand{\D}{\mathcal{D}}
\newcommand{\Dc}{\mathcal{D}_\text{ctx}}
\newcommand{\Dq}{\mathcal{D}_\text{qry}}

\makeatletter
\newcommand\footnoteref[1]{\protected@xdef\@thefnmark{\ref{#1}}\@footnotemark}
\makeatother

\newcommand{\cbit}{\begin{compactitem}}
\newcommand{\ceit}{\end{compactitem}}
\newcommand{\cben}{\begin{compactenum}}
\newcommand{\ceen}{\end{compactenum}}

\newcommand{\beq}{\begin{equation}}
	\newcommand{\eeq}{\end{equation}}

\newcommand{\fomo}{{\sc FoMo-0D}\xspace}

\newcommand{\outformer}{{\sc OutFormer}\xspace}
\newcommand{\tactic}{{\sc TACTIC}\xspace}
\newcommand{\iclad}{{\sc ICLAD}\xspace}

\definecolor{darkgreen}{RGB}{41,166,41}

\newcommand{\bit}{\begin{itemize}}
	\newcommand{\eit}{\end{itemize}}
\newcommand{\ben}{\begin{enumerate}}
	\newcommand{\een}{\end{enumerate}}

\newcounter{x}

\newcommand{\bx}{\mathbf{x}}

\newcommand{\bp}{\mathbf{p}}

\definecolor{celadon}{rgb}{0.67, 0.88, 0.69}
\definecolor{carolinablue}{rgb}{0.6, 0.73, 0.89}

\definecolor{aliceblue}{rgb}{0.867, 0.917, 0.964}
\definecolor{aliceyellow}{rgb}{0.999, 0.945, 0.796}
\definecolor{alicegray}{rgb}{0.844, 0.867, 0.898}

\usepackage{xspace}
\usepackage{threeparttable}
\usepackage{hyperref}
\usepackage{url}
\usepackage{enumitem}
\usepackage{graphicx}
\usepackage{wrapfig}
\usepackage{placeins}
\usepackage{float}
\usepackage{booktabs}
\usepackage{amssymb}
\usepackage{pifont}
\usepackage{multirow}
\usepackage{subcaption}
\usepackage{tabularx}
\usepackage{bm}

\title{{When Less Compute Is More: Adaptive Early Exit Improves %
Pretrained Outlier Detection}}

\author{Tianyang Zhou \\
Carnegie Mellon University \\
\texttt{tzhou3@andrew.cmu.edu} \\
\And
Leman Akoglu \\
Carnegie Mellon University \\
\texttt{lakoglu@andrew.cmu.edu}
}

\iclrfinalcopy %
\begin{document}

\maketitle
\lhead{Preprint}

\begin{abstract}
Pretrained tabular foundation models process every dataset at a fixed depth, with inference costs growing with dataset size. To address this, we present the first study of depth-adaptive early-exit for pretrained outlier detection models. While early-exit is typically motivated by efficiency, we uncover a surprising benefit: exiting at the optimal intermediate layer \textit{can also improve detection performance} on diverse real-world benchmarks by 4.7--7.3\% on average, consistent across three distinct foundation models.
First, we investigate the factors driving these gains, and identify a key mechanism: \textit{context pollution}, i.e., the presence of outliers among in-context samples. Our analysis reveals that nearby in-context samples exert increasing influence on query predictions at greater depths, consistent with a retrieval-based view of these models.
In effect, early-exit alleviates the adverse effects of retrieving accurate-yet-polluted neighbors, with gains of \mbox{13--21\%} when context pollution matches the natural outlier rate.
Motivated by these findings, we pretrain a plug-in router  to select a dataset-specific exit layer, using query outlier labels as privileged information available only during router training. The router operates \textit{post hoc}, leaving the base  model parameters and prediction head unchanged.
Experiments on three large real-world benchmarks %
 show that, on clean context, the router recovers up to 45\% of the oracle gain with up to 1.8$\times$ speedup across three pretrained backbones, with larger gains as context pollution increases.

\end{abstract}

\vspace{-.1in}
\section{Introduction}
\vspace{-.05in}
\label{sec:intro}

Empirical neural scaling laws \citep{kaplan2020scaling} have motivated the  rapid scaling of foundation models in both parameter capacity and depth, often reaching billions of parameters \citep{hoffmann2022training}.
However, increasing model sizes also raise the computational cost of inference, even for inputs that may not require the model's full capacity. 
This motivates \emph{compute-adaptive inference}: allocating computational effort according to the given input and task~\citep{schwartz2020right}.

Adaptive computation can be introduced through architectural design, training for variable-depth execution, or post-hoc adaptation of pretrained backbones.
Natively adaptive designs  include Looped Transformers that vary effective depth through repeated block execution \citep{giannou2023looped,geiping2025scaling} and Mixture-of-Experts that route tokens to specialized sub-networks to decouple massive capacity from active per-token compute \citep{shazeer2017outrageously,Fedus2021SwitchTS,glamgoogle}.
Beyond architectural designs, other work trains auxiliary adapters alongside the backbone without altering the core architecture to enable dynamic early-termination, such as entropy-based exits \citep{xin2020deebert,liu2020fastbert} and layer-specific prediction heads \citep{zhou2020pabee}.

The cost of retraining large adaptive models motivates a second line of work that trains only auxiliary modules. These approaches retrofit existing pretrained models to reduce inference depth by introducing lightweight decision mechanisms  such as confidence-based stopping criteria \citep{schuster2022confident,kuken2025early}, frozen-backbone intermediate decoders \citep{kuken2025early}, and learned post-hoc routing policies \citep{he2025router,luo2025adaptive}. (See Appdx. \ref{sec:related} for details.)

In this work, we investigate \emph{post hoc depth-adaptive early exit} for pretrained tabular outlier detection models. These models process every dataset via in-context learning at a fixed depth irrespective of the input. We seek to terminate inference dynamically at dataset-specific intermediate layers, reserving deeper processing for datasets that benefit from it.

\begin{figure}[!t]
\vspace{-.2in}
\hspace{-.2in}
\centering
\includegraphics[width=\linewidth]{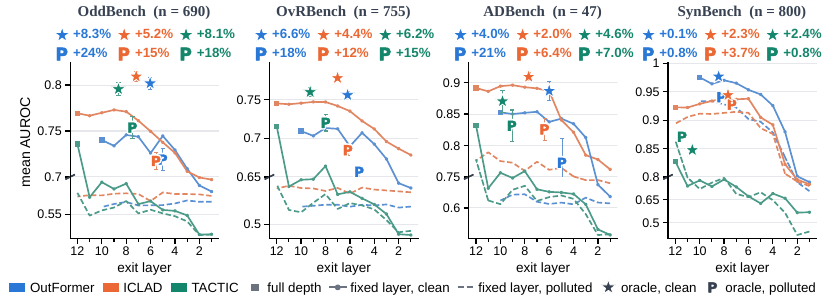}
\caption{OD performance when exiting at a fixed layer under clean (--) and polluted (- -) context; context pollution hinders performance across different pretrained models and benchmarks. Per-dataset oracle early-exit boosts performance for both clean ($\bigstar$) and especially polluted (\textbf{P}) context.}
\label{fig:layers}
\end{figure}

While adaptive inference traditionally targets efficiency, 
our investigation uncovers a surprising opportunity: reducing depth can improve both latency \textit{and} detection performance.
We trace a key part of these gains to \emph{context pollution}, which causes outlier masking: deeper processing strengthens the influence of nearby context samples, which can harm predictions when those neighbors are themselves outliers.
{Early exit helps mitigate masking by limiting the accumulation of context pollution in query representations, making early-exit gains under pollution more pronounced (see Figure \ref{fig:layers}).}

Motivated by these dual benefits, we pretrain a plug-in router, called \method, to select a dataset-specific exit layer, leaving the base foundation model and its prediction head entirely unchanged. A core challenge is that while context outliers amplify early-exit benefits, inference-time datasets lack ground truth labels. To bridge this gap, we leverage query outlier labels as privileged information \citep{vapnik2015learning} exclusively during \method's training, requiring no labels at inference. 
Our main contributions are summarized as follows.

\begin{itemize}
[leftmargin=10pt, itemsep=-0.01in, topsep=-0.025in]

\item \textbf{Post-hoc Early Exit for Pretrained Outlier Detection. }We present the first study of depth-adaptive early exit for pretrained outlier detection (OD) models, addressing dataset-specific exit selection without labeled outliers in target datasets. Our approach preserves the pretrained backbone and prediction head, requiring only a lightweight routing module.

\item \textbf{``Pollution in the Deep'': Why Early Exit Can Improve Detection.}
We uncover a dual benefit of early exit: reducing computation can also improve detection performance. We identify context pollution as a key mechanism, showing that deeper layers increasingly rely on nearby in-context samples whose influence can become detrimental when those neighbors are outliers. Early exit alleviates this effect: 
across three distinct pretrained  OD backbones, optimal intermediate layer predictions  outperform their full-depth counterparts on diverse real-world benchmarks by \mbox{4.7--7.3\%} on average, while also reducing computation.

\item \textbf{Learning to Route under Context Pollution without Inference-time Labels.}
We introduce \method, a dataset-adaptive router 
that combines sequential stopping and retrospective exit selection with annealed query labels as privileged information during synthetic pretraining, preserving the backbone and prediction head and requiring no inference-time labels.

\item \textbf{Evaluation across Backbones, Benchmarks, and Pollution Levels.}
Experiments across three pretrained backbones, \outformer \citep{ding2026zero}, \tactic \citep{marszalek2026tactic}, and \iclad \citep{wei2026iclad}, and three large real-world benchmarks and a synthetic benchmark \citep{ding2026macrodata,han2022adbench} establish that
higher accuracy and lower latency at  inference
are not mutually exclusive objectives for pretrained outlier detection models, where strategic early-exit by \method yields up to 45\% of the oracle gain and up to 1.8$\times$ inference speedups on clean context, with larger gains as context pollution increases.

\end{itemize}

\vspace{-.1in}
\section{Background and Findings}
\vspace{-.05in}
\label{sec:prelim}

\textbf{Preliminaries.} Pretrained tabular foundation models (TFMs) address the learning problem via \textit{in-context} learning \citep{xie2022explanation}. A given dataset $\D$, comprising samples in $\R^d$, is partitioned into two parts: a set of context points $\Dc \subset \D$ representing the training data, and a query set $\Dq = \D \setminus \Dc$ serving as the unlabeled test set. In supervised TFMs for standard classification or regression, $\Dc$ is labeled, with feature vectors concatenated with their respective target labels, whereas outlier detection TFMs are pretrained using both unlabeled context and query sets.

TFMs directly model the posterior predictive distribution $p(y \mid \bx, \Dc)$ over the  unlabeled query set $\Dq$ given context set $\Dc$. They  are typically optimized on synthetic datasets $\D \sim \pi(\D)$ drawn from a data-generating distribution $\pi(\cdot)$, specifying prior probabilities over datasets. Training is supervised using cross-entropy loss to predict the labels of query points given the context. Pretrained outlier detection models specifically leverage ground-truth outlier/inlier labels from the synthetic priors for supervision, while context points are provided without labels. The training objective is 
\begin{equation}
    \theta^\star = \arg\min_{\theta} \mathbb{E}_{\D \sim \pi(\D)} \big[ \sum_{\bx \in \Dq} \ell\left(y(\bx), \hat{p}_\theta(y = \text{outlier} \mid \bx, \Dc)\right) \big] \;.
\end{equation}
During zero-shot inference, a pretrained TFM labels outliers in input datasets in a single forward pass by conditioning directly on an unlabeled context set  without any updates to the model weights.

\textbf{Outlier Detection TFMs.}
Pretrained tabular outlier detection (OD) models differ along two primary axes: (1) their synthetic data priors over inlier distributions and outlier-generating mechanisms, and (2) their training supervision regimes. 
\fomo \citep{shen2025fomod}, the seminal TFM for zero-shot OD, is pretrained on Gaussian mixture models (GMMs) featuring subspace outliers with inflated variance across random subsets of features, operating exclusively under the \textbf{one-class, clean context} setting with inlier-only context points. 
Its successor, \outformer \citep{ding2026zero}, broadens this paradigm by training on a mixture of heterogeneous priors, including GMMs, structural causal models (SCMs), and copulas, while maintaining an inlier-only clean context.

Similarly, \tactic \citep{marszalek2026tactic} leverages a hybrid prior of SCMs and GMMs, modeling rare-class outliers from SCMs alongside local, global, and clustered outliers derived from variance-inflated or mean-shifted GMMs. 
While \tactic-C adopts the standard  one-class setting, \tactic-N addresses the  \textbf{unsupervised, polluted context} scenario by allowing unlabeled outliers within the context to enhance real-world robustness. 
Finally, \iclad \citep{wei2026iclad} employs SCM-based datasets incorporating rare-class outliers and additive-noise subspace outliers, while expanding supervision  to encompass one-class, unsupervised, as well as {semi-supervised} context where a small subset of context points receive explicit outlier labels while the rest stay unlabeled.

\begin{wrapfigure}{r}{0.40\linewidth}
\vspace{-0.08in}
\hspace{-0.25in}
\centering
\includegraphics[width=1.1\linewidth]{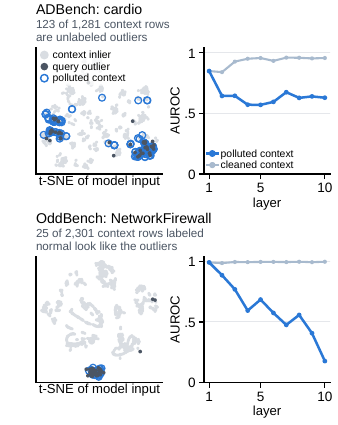}
\vspace{-0.15in}
\caption{Hidden pollution in two official real-world splits: (left) t-SNE of the model input, polluting context rows circled; (right) layer-wise AUROC (\outformer) with and without them.}
\label{fig:unknown_pollution}
\vspace{-0.05in}
\end{wrapfigure}
\textbf{Early Exiting OD-TFMs -- Setup.}
\textit{Backbones.} We study three OD-TFMs with distinct training regimes (\outformer, \tactic, and \iclad) to analyze their early-exit behavior across two context settings: \textit{inlier-only}  and \textit{polluted}. Context points are provided without labels, i.e., the semi-supervised regime is excluded. Each backbone is kept entirely frozen, including the parameters of its final classification head. For prediction, representations of query points from intermediate layers are routed to the fixed head to quantify detection performance. The dataset-specific optimal layer yielding peak performance is designated as the \textit{oracle layer}.

\vspace{-.05in}
\textit{Datasets.} Our study spans 
(1) {SynBench} with 800 synthetic datasets generated from \outformer's priors (GMMs, SCMs,  copulas); 
(2) {ADBench} \citep{han2022adbench}, comprising its 47 classical tabular OD datasets; and 
(3) {OddBench} and (4) {OvRBench}, both sourced from MacrOData \citep{ding2026macrodata}, OddBench featuring 690 datasets with real-world anomalies (system faults, errors, malware, etc.) and OvRBench consisting of 755 repurposed classification datasets, one class designated as inliers and the rest subsampled to form outliers.

\vspace{-.05in}
\textit{Context Pollution.} While SynBench supplies perfectly clean, simulated inlier-only context data via its standard splits, real-world datasets may provide unsupervised random splits, or may harbor inherent label noise, which we refer to as ``hidden pollution'', reflecting the
difficulty of acquiring purely inlier data in practice. Figure \ref{fig:unknown_pollution} illustrates these issues on two real-world datasets. We study context pollution by matching the proportion of outliers in context and query sets to the natural rate in the underlying dataset.  %

\begin{figure}[!t]
\vspace{-0.2in}
\centering
\includegraphics[width=\linewidth]{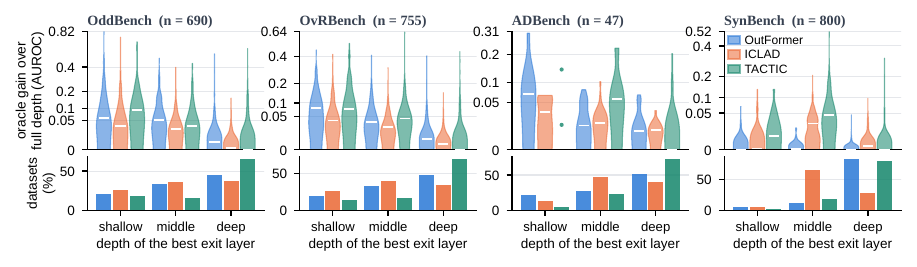}
\vspace{-0.2in}
\caption{Oracle gain over full depth, grouped by which third of the depth the best exit layer falls in (square-root scaled y-axis); bars give the share of datasets per group.}
\label{fig:gain_depth}
\vspace{-0.05in}
\end{figure}

\textbf{Early-Exit Results \& Observations.}
Figure~\ref{fig:layers} illustrates detection performance across layers when each model exits at a \textit{fixed layer} $l$$\in$$\{\text{\tt max\_depth}, \ldots, 1\}$ under both clean and {polluted (at original rate) context}, 
where \texttt{max\_depth} is 10 for \outformer and 12 for both \tactic and \iclad. 
Context pollution consistently degrades detection performance across all models. %
Average performance across datasets remains fairly stable and comparable to the final-layer performance before it starts a considerable decline. %
Surprisingly, when models exit at the dataset-specific \textit{oracle layer}, average performance actually \textit{improves} relative to the final layer for all models, as depicted with  $\bigstar$ (7.3\% for \outformer, 7.0\% for \tactic, and 4.7\% for \iclad across real-world benchmarks). Performance gains are even \textit{more pronounced under polluted context}, as depicted with \textbf{P} (20.9\% for \outformer, 16.3\% for \tactic, and 13.2\% for \iclad across real-world benchmarks). 
This lift in gains from clean to polluted is significant across models and benchmarks, as Table \ref{tab:pvals} shows.
\begin{table*}[h]
\centering
\caption{Oracle performance gains (\%): Clean $\to$ Polluted; $p$-values in parentheses.}
\vspace{-0.05in}
\small
\setlength{\tabcolsep}{4pt}
\setlength{\aboverulesep}{1pt}
\setlength{\belowrulesep}{1pt}
\renewcommand{\arraystretch}{0.85}
\begin{tabular}{@{}l*{3}{r@{\hspace{3pt}$\to$\hspace{5pt}}l}@{}}
\toprule
\rule[-3.5pt]{0pt}{12pt}Backbone & \multicolumn{2}{c}{OddBench (690)}
         & \multicolumn{2}{c}{OvRBench (755)}
         & \multicolumn{2}{c}{ADBench (47)} \\
\midrule
OutFormer
& $+8.3$ & $+24.0\;(3{\times}10^{-48})$
& $+6.6$ & $+18.1\;(2{\times}10^{-56})$
& $+4.0$ & $+21.4\;(2{\times}10^{-8})$ \\
TACTIC
& $+8.1$ & $+18.5\;(3{\times}10^{-48})$
& $+6.2$ & $+15.0\;(2{\times}10^{-44})$
& $+4.6$ & $+7.0\;(0.086)$ \\
ICLAD
& $+5.2$ & $+15.4\;(4{\times}10^{-60})$
& $+4.4$ & $+11.6\;(2{\times}10^{-48})$
& $+2.0$ & $+6.4\;(10^{-5})$ \\
\bottomrule
\end{tabular}
\label{tab:pvals}
\vspace{-7pt}
\end{table*}

Figure  \ref{fig:gain_depth} (top) shows the performance gain distribution under {clean context} by oracle depth, grouped as shallow, middle and deep. 
While relative gains are {marginal on the  synthetic SynBench}  with perfectly clean context,  they become \textit{more pronounced on real-world} datasets (0.1\% vs 7.3\% for \outformer, 2.4\% vs 7.0\% for \tactic, and 2.3\% vs 4.7\% for \iclad). 

Figure \ref{fig:gain_depth} (bottom) displays the share of datasets whose oracle layer falls in each depth group %
(per-layer histograms in Figure~\ref{fig:exit_dist}). 
Oracle layers vary notably for all models across all benchmarks, underscoring the potential gains that early exit offers for OD-TFMs. 
A notable observation for \outformer is that  the majority of SynBench datasets accumulate at the final layer as their oracle layer, reflecting the model's pretraining on %
matching underlying priors for final-layer optimization. In contrast, the distribution disperses considerably across real-world benchmarks.

\vspace{-.1in}
\section{The Paradox of Early Exit: Understanding Performance Gains}
\vspace{-.05in}
\label{sec:understanding}

Before translating these findings into a practical early-exit strategy for OD-TFMs, we seek to understand the apparent paradox: intermediate representations can yield substantially better predictions through a head trained exclusively on final-layer outputs, without updating any parameters.

The decline in detection performance and increase in oracle early-exit gains as we move from purely clean synthetic data to real-world benchmarks, and further to polluted contexts, suggest that in-context outliers play a crucial role. 
Real-world gains under nominally clean contexts may likewise stem from ``hidden pollution'', i.e., outlier-like samples within designated inlier sets.

In-context learning has been interpreted through several perspectives, including implicit Bayesian inference \citep{xie2022explanation}, implicit gradient descent \citep{dai2023whycangpt}, and specialized circuits such as induction heads \citep{olsson2022incontext}. Particularly relevant to our observations is the retrieval-based perspective proposed for tabular generalization, whereby models identify and aggregate relevant context examples \citep{shaheen2026understanding}. This view is consistent with connections between attention, kernel smoothing, and non-parametric regression \citep{ilin2026discoformer,santos2026sparse}.

\begin{figure}[th]
\vspace{-0.1in}
\centering
\includegraphics[width=\linewidth]{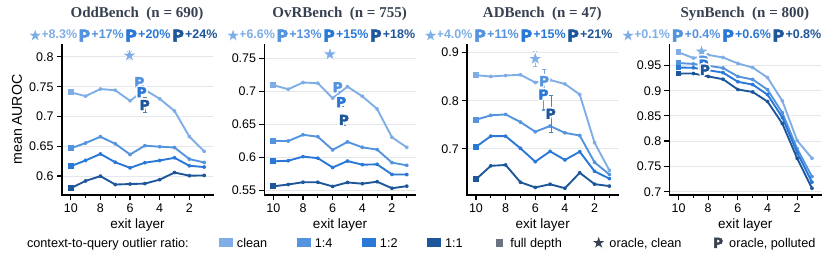}
\vspace{-0.15in}
\caption{Fixed-layer and oracle performance of \outformer as held-out context pollution increases; $\bigstar$ depicts the oracle on clean context and \textbf{P} the oracle under pollution.
}
\label{fig:pollution_dose}
\end{figure}

Under this interpretation, nearby context outliers may provide misleading evidence for a query outlier, inducing ``\textbf{outlier masking}'': outliers in training data make similar query outliers appear less anomalous. We hypothesize that repeated context aggregation across layers amplifies this masking effect, progressively suppressing the anomaly scores of affected queries. 
Early exit may thus mitigate outlier masking by limiting the accumulation of context pollution in query representations.

We probe this hypothesis in two pollution setups. In \textbf{held-out pollution}, we start from clean context and   progressively inject known held-out outliers, adjusting the context-to-query outlier ratio across 1:4, 1:2, and 1:1 (natural rate). Figure~\ref{fig:pollution_dose} shows for \outformer that increasing pollution degrades detection performance while amplifying the gains from oracle early-exit, consistent with earlier observations (similar results for \iclad and \tactic are in Figs.~\ref{fig:pollution_dose_fixedhalf_iclad} and \ref{fig:pollution_dose_fixedhalf_tactic}).
We also inject \textbf{near-duplicate pollution}, where we plant ``sibling'' outliers into the context by creating perturbed copies of the query outliers. The results strongly align with prior observations. (See Figs.~\ref{fig:pollution_dose_sibling}, \ref{fig:pollution_dose_iclad}, and \ref{fig:pollution_dose_tactic}.)

\begin{figure}[h]
\vspace{-0.015in}
\centering
\includegraphics[width=0.92\linewidth]{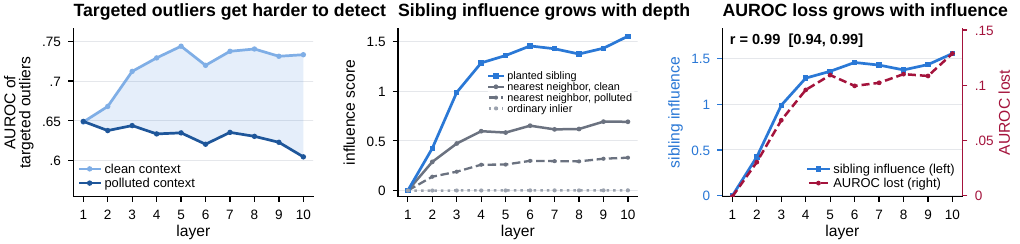}
\vspace{-0.05in}
\caption{Effect of one planted sibling per targeted outlier (\outformer): \textbf{(left)} Detection of targeted outliers declines with depth; \textbf{(middle)} Outlier scores increasingly rely on the sibling with depth; \textbf{(right)} Sibling influence and AUROC loss rise together, showing a strong correlation of 0.99.}
\label{fig:gradient}
\vspace{-0.015in}
\end{figure}

Next, we quantify \textbf{sibling influence} at each exit layer using gradient-based saliency~\citep{kokhlikyan2020captum}: the gradient norm of a query's outlier score with respect to its sibling's input features.
We normalize this norm for scale differences across datasets (details in Appdx.~\ref{ssec:influence}).
Figure~\ref{fig:gradient} presents results for \outformer (see also Figs.~\ref{fig:gradient_iclad} and \ref{fig:gradient_tactic}). Notably, targeted pollution increasingly impairs detection (left). Furthermore, sibling influence, and thus the impact of pollution, increases with depth (middle); in fact, the performance loss closely tracks this influence across layers (right).

\vspace{-.1in}
\section{Layer-Adaptive Routing: Learning When to Exit}
\vspace{-.05in}
\label{sec:method}
\begin{figure}[!h]
\vspace{-0.1in}
\centering
\includegraphics[width=\linewidth]{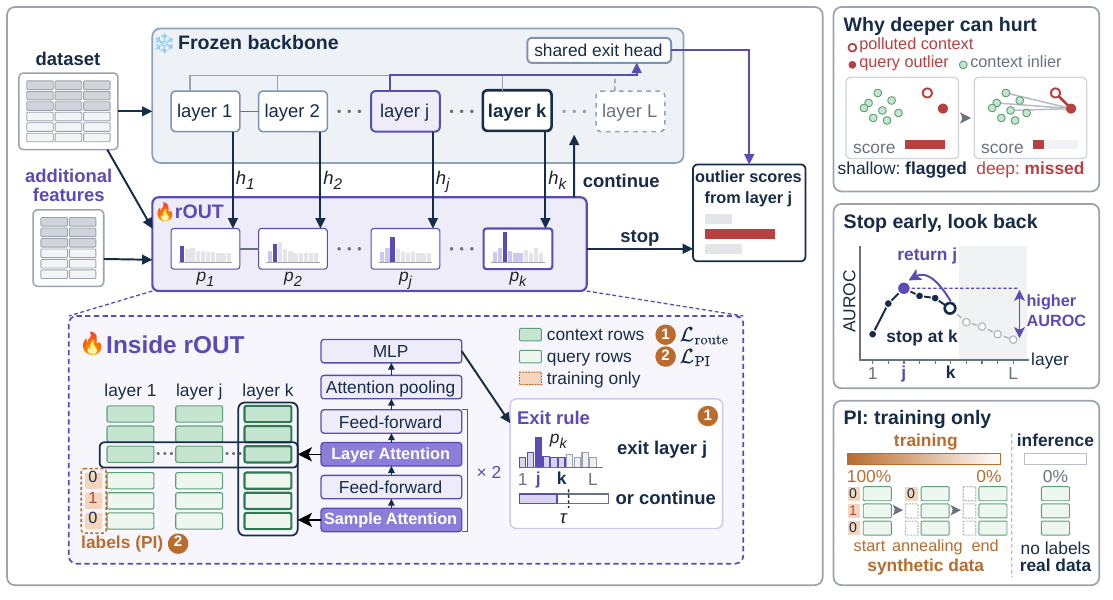}
\vspace{-0.2in}
\caption{\textbf{Left:} \method reads the frozen backbone layer by layer, stops at layer $k$, and returns the best computed layer $j\le k$. \textbf{Right:} deeper layers can lean on polluted context that masks query outliers; thus, \method stops early and looks back; query labels are training-only privileged information.
}
\label{fig:arch}
\end{figure}

Many depth-adaptive methods use backbone activations to route computation layer by layer~\citep{elbayad2020depth,heakl2026drllm,kuken2025early}. They do not separately choose among predictions from layers already computed. For OD-TFMs, this matters: deeper layers can \textit{impair} detection;
thus, getting stuck in a later layer without \textbf{retrospective recovery} can be detrimental. To enable this,  \method introduces \textbf{layer-wise attention} to separate computation depth from output selection, comparing all computed layers to decide when to stop and which prediction to return.

\textbf{Architecture.~}
At its core, \method uses layer-wise context and query representations to select an exit layer. As Figure~\ref{fig:arch} shows, it comprises two attention modules: Sample Attention  attends across samples at layer $k$, while Layer Attention  attends across each sample's representations from layers $1$ through $k$. Following learned attention pooling~\citep{lee2019set}, an MLP predicts a distribution $\bp_k$ over all $L$ candidate exit layers. We stop at the first layer where the probability mass on the layers already computed reaches $\tau$, i.e., $K=\min\{k:\sum_{j\le k}p_{k,j}\ge\tau\}$.
At that point, \method returns the backbone's predictions from the most probable layer among those already computed, $\hat{\jmath}=\arg\max_{j\le K}p_{K,j}$. Following the validation-based threshold selection of \citet{elbayad2020depth}, we select a single $\tau$ per backbone on synthetic validation datasets to maximize $\mathbb{E}_{\mathrm{val}}[\mathrm{AUROC}_{\hat{\jmath}}-\lambda K]$, where $\lambda$ controls the depth penalty. We apply this threshold unchanged to every test dataset.

\textbf{Training Objective.~}
We train \method to balance the quality of the returned predictions against the depth computed. For each training dataset, let $A_j$ be the backbone's AUROC at exit layer $j$, computed with the query outlier labels, and $R_j=\max_i A_i-A_j$ the regret of selecting layer $j$. The depth penalty $\lambda$ controls the trade-off between AUROC and computation: without it, the router tends to defer its decision until the final few layers, resulting in little computational saving (see Table~\ref{tab:lambda}).

\vspace{-0.025in}
\textit{Sequential stopping loss.~}
We formulate sequential stopping probabilistically and optimize the expected loss over stopping depths~\citep{banino2021pondernet}.  Conditioned on reaching layer $k$, the stopping probability (as a function of trainable router parameters $\theta$)  is $\pi_k(\theta)=\sum_{j\le k}p_{k,j}(\theta)$, giving $w_k(\theta)=\pi_k(\theta)\prod_{i<k}(1$$-$$\pi_i(\theta))$ as the probability of stopping at $k$, with $\sum_{k=1}^{L}w_k(\theta)=1$. Then, %

\vspace{-0.2in}
\begin{equation}
\mathcal{L}_{\mathrm{seq}}(\theta)=\sum_{k=1}^{L}w_k(\theta)\Big(\sum_{j\le k}\frac{p_{k,j}(\theta)}{\pi_k(\theta)}R_j+\lambda k\Big),
\end{equation}
where the first term measures the {expected} regret of the layer returned upon stopping at $k$, which may be an earlier layer $j<k$, while the second penalizes the computation required to reach $k$. Training with a cross-entropy loss that targets the best layer would penalize all other exits equally, whereas the regret penalizes each exit in proportion to its AUROC gap from the best layer.

\vspace{-0.025in}
\textit{Layer-wise selection loss.~}
To provide a direct exit-selection signal at every depth, we additionally supervise $\bp_k(\theta)$ independently of the stopping probabilities:

\vspace{-0.15in}
\begin{equation}
\mathcal{L}_{\mathrm{layer}}(\theta)=\frac{1}{L}\sum_{k=1}^{L}\sum_{j=1}^{L}p_{k,j}(\theta)\big(R_j+\lambda j\big).
\end{equation}
\vspace{-0.15in}

$\mathcal{L}_{\mathrm{layer}}$ incentivizes $\bp_k$ to assign high probability to layers with a favorable accuracy--computation trade-off. Here each candidate is penalized for its own depth $j$; in contrast, $\mathcal{L}_{\mathrm{seq}}$ penalizes the depth $k$ actually computed, which exceeds $j$ when the router returns an earlier layer.

The two terms form the routing objective, $\mathcal{L}_{\mathrm{route}}=\mathcal{L}_{\mathrm{seq}}+\alpha\mathcal{L}_{\mathrm{layer}}$, with weight $\alpha$ on layer-wise selection. We additionally use entropy regularization to prevent the selection distributions from collapsing onto a single layer (Appdx.~\ref{app:objective}).

\textbf{Outlier Labels as Privileged Information (PI).~}
We pretrain \method on synthetic datasets drawn from diverse data priors, with ground-truth outlier labels and varying levels of context pollution. Query outlier labels are available during pretraining, whereas inference-time datasets are fully unlabeled. Following the learning using privileged information (LUPI) framework \citep{vapnik2009new,vapnik2015learning}, we use these labels both as supervision and as privileged inputs.

\vspace{-0.1in}
\textit{PI as supervision.}
$\mathcal{L}_{\mathrm{route}}$ uses query labels through \textit{overall} layer-wise AUROC. We also use them to directly supervise the router's representations through an auxiliary objective $\mathcal{L}_{\mathrm{PI}}$, including binary cross-entropy for {predicting \textit{individual} query rows' outlier labels}. This supervision helps the router distinguish inliers from outliers when selecting an exit. We minimize the combined loss %
\begin{equation}
\min_{\theta} \;\; \mathcal{L}_{\mathrm{route}}(\theta)+\mathcal{L}_{\mathrm{PI}}(\theta)\;.
\end{equation}
The auxiliary terms and their weights are detailed in Appdx.~\ref{app:objective}.

\textit{PI as annealed input.~}
Besides outlier label supervision, we expose outlier structure through {privileged inputs}: learned embeddings of the revealed query labels and label-derived summaries of the query outliers. Training begins with all query labels and their summaries available, then progressively withdraws both, using cosine schedules (Appdx.~\ref{app:objective}) so that \method learns to route without them, as PI input is only available during training. When a query label is withheld, its embedding is replaced by a shared embedding indicating that the label is unavailable. The final training phase uses no privileged inputs, matching the fully unlabeled inference setting.

\vspace{-.015in}
\textbf{Additional Features.~}
Alongside the backbone representations, the router receives, for each row of the dataset, its raw input features, the backbone's layer-wise outlier scores, and 29 geometric features (Appdx. \ref{app:features}). The geometric features describe neighborhood distances, local density, and each row's position relative to the context distribution. Raw inputs and scores from intermediate prediction heads add little computational overhead, and we restrict the types and number of geometric features to keep feature extraction overhead low.

\vspace{-.015in}
\textbf{Pretraining Priors.~}
Finally, motivated by \iclad's %
extensive priors, we extend \outformer's data priors in two ways. First, we add unsupervised detection tasks with polluted contexts, including near-duplicate pollution. Second, we complement its %
mixed priors with class-structured outliers,  by relabeling classes of TabPFN's %
classification datasets (Table~\ref{tab:corpus}).

\vspace{-.1in}
\section{Experiments}
\vspace{-.075in}
\label{sec:exp}

\textbf{Datasets and evaluation settings.}
We evaluate \method on ADBench~\citep{han2022adbench}, OddBench, and OvRBench~\citep{ding2026macrodata}, three real-world outlier detection benchmarks comprising 47, 690, and 755 datasets, respectively, and on SynBench~\citep{ding2026macrodata}, a synthetic benchmark comprising 800 datasets (Table~\ref{tab:benchmarks}).
We evaluate under clean context as well as the held-out and near-duplicate pollution settings described in Section~\ref{sec:understanding}.

\vspace{-.015in}
\textbf{Baselines.}
Besides \textbf{Full depth}, we consider \textbf{Half depth} as a simple heuristic baseline. We also compare to the tabular early-stopping approach of \citet{kuken2025early}, which we call \textbf{LA-Entropy}; it trains separate layer-specific prediction heads (decoders) per model post hoc and exits when the entropy of the predictive distribution falls below a threshold $\tau$, returning the corresponding predictions. Following the original formulation, we use a single $\tau$ shared across all exit heads. We tune $\tau$ on synthetic validation datasets and keep it fixed across all test datasets. (See Figure~\ref{fig:tau_sweep} for the full threshold sweep.)
We also report \textit{Oracle}, depicting peak performance at the dataset-specific optimal exit layer, and \textit{Best fixed layer}, selecting the same layer for all datasets within a benchmark to maximize average AUROC. Both use ground-truth test AUROCs and are reported as references.

\vspace{-.015in}
\textbf{Metrics.}
We report mean AUROC and the mean number of computed layers, counting all layers evaluated before the exit decision, even if \method selects an earlier layer for prediction. We report the relative AUROC gain of $m\in\{\text{\method},\text{oracle}\}$ over full depth, $\frac{\bar{A}_{m}-\bar{A}_{\mathrm{full}}}{\bar{A}_{\mathrm{full}}}\times 100\%$, where $\bar{A}$ is mean AUROC across a benchmark's datasets. In Table~\ref{tab:main}, parentheses beside \method's AUROC give the share of the full-depth-to-oracle gap it closes, $\frac{\bar{A}_{\text{\method}}-\bar{A}_{\mathrm{full}}}{\bar{A}_{\mathrm{oracle}}-\bar{A}_{\mathrm{full}}}\times 100\%$. For each benchmark, we report total computation in PFLOPs and total wall-clock inference time in seconds across all datasets on one H100 GPU, including backbone computation, router calls, and additional-feature computation.

\vspace{-.015in}
\textbf{Implementation details.}
We use the same router architecture for all three backbones: two blocks of Sample and Layer Attention with hidden size 256 and four attention heads, followed by learned attention pooling, totaling 4.6M parameters per router. The router processes up to 256 context and 1,024 query rows per dataset, while the backbone scores all query rows. We train on over 200,000 synthetic datasets and use around 6,000 held-out datasets for validation. Training uses AdamW and the privileged-input schedule described in Section~\ref{sec:method}. At inference, we average the routing logits of three independently trained seeds per backbone. The stopping threshold is selected on synthetic validation data. Full training and hyperparameter settings are provided in Appdx.~\ref{app:objective}.

\begin{table}[!t]
\vspace{-.1in}
\centering
\caption{\method against  fixed-depth baselines and per-layer optimized LA-Entropy{$^\triangle$} \citep{kuken2025early} across benchmarks and backbones.  %
\textbf{Bold} is the best deployable method  (\textit{Oracle} and \textit{Best fixed layer} are for reference).
In parentheses is the share of Full-depth-to-\textit{Oracle} gap that \method closes. %
}
\label{tab:main}
{\footnotesize\setlength{\tabcolsep}{2.2pt}
\ifdefined\tabmainbox\else\newsavebox\tabmainbox\fi
\sbox\tabmainbox{\renewcommand{\arraystretch}{0.923}%
\begin{tabular*}{420pt}{@{}cl@{\extracolsep{\fill}}cccccccc@{}}
\toprule
 &  & \multicolumn{2}{c}{\small OddBench} & \multicolumn{2}{c}{\small OvRBench} & \multicolumn{2}{c}{\small ADBench} & \multicolumn{2}{c}{\small SynBench} \\
\cmidrule(lr){3-4}\cmidrule(lr){5-6}\cmidrule(lr){7-8}\cmidrule(lr){9-10}
Backbone & Method & AUROC & Layers & AUROC & Layers & AUROC & Layers & AUROC & Layers \\
\midrule
\multirow{6}{*}{\shortstack{OutFormer\\ ($L{=}10$)}} & \textit{Oracle} & 0.802 & 6.0 & 0.756 & 6.2 & 0.887 & 6.0 & 0.977 & 8.4 \\
 & \textit{Best fixed layer} & 0.746 & 8.0 & 0.713 & 8.0 & 0.854 & 7.0 & 0.975 & 10.0 \\
\cline{2-10}\noalign{\smallskip}
 & Full depth & 0.740 & 10.0 & 0.709 & 10.0 & 0.853 & 10.0 & \textbf{0.975} & 10.0 \\
 & Half depth & 0.745 & 5.0 & 0.707 & 5.0 & 0.843 & 5.0 & 0.945 & 5.0 \\
 & LA-Entropy & 0.709 & 2.5 & 0.680 & 2.9 & 0.770 & 2.5 & 0.915 & 2.6 \\
 & \textbf{rOUT (ours)} & \textbf{0.761} \scriptsize{(34\%)} & 5.5 & \textbf{0.730} \scriptsize{(45\%)} & 5.4 & \textbf{0.860} \scriptsize{(21\%)} & 5.5 & 0.967 & 5.7 \\
\midrule
\multirow{6}{*}{\shortstack{ICLAD\\ ($L{=}12$)}} & \textit{Oracle} & 0.809 & 7.2 & 0.778 & 7.0 & 0.909 & 7.7 & 0.944 & 7.7 \\
 & \textit{Best fixed layer} & 0.773 & 9.0 & 0.747 & 9.0 & 0.896 & 9.0 & 0.939 & 8.0 \\
\cline{2-10}\noalign{\smallskip}
 & Full depth & 0.769 & 12.0 & 0.745 & 12.0 & 0.891 & 12.0 & 0.923 & 12.0 \\
 & Half depth & 0.750 & 6.0 & 0.735 & 6.0 & 0.887 & 6.0 & 0.938 & 6.0 \\
 & LA-Entropy & 0.704 & 1.6 & 0.696 & 2.0 & 0.786 & 2.1 & 0.856 & 1.8 \\
 & \textbf{rOUT (ours)} & \textbf{0.776} \scriptsize{(18\%)} & 6.8 & \textbf{0.755} \scriptsize{(30\%)} & 6.0 & \textbf{0.891} \scriptsize{(0\%)} & 6.0 & \textbf{0.938} \scriptsize{(71\%)} & 4.4 \\
\midrule
\multirow{6}{*}{\shortstack{TACTIC\\ ($L{=}12$)}} & \textit{Oracle} & 0.796 & 8.6 & 0.760 & 9.3 & 0.870 & 9.8 & 0.847 & 10.6 \\
 & \textit{Best fixed layer} & 0.736 & 12.0 & 0.715 & 12.0 & 0.832 & 12.0 & 0.827 & 12.0 \\
\cline{2-10}\noalign{\smallskip}
 & Full depth & 0.736 & 12.0 & 0.715 & 12.0 & 0.832 & 12.0 & 0.827 & 12.0 \\
 & Half depth & 0.603 & 6.0 & 0.603 & 6.0 & 0.678 & 6.0 & 0.673 & 6.0 \\
 & LA-Entropy & 0.693 & 2.1 & 0.682 & 2.4 & 0.780 & 2.2 & \textbf{0.876}\textsuperscript{$\ast$} & 1.9 \\
 & \textbf{rOUT (ours)} & \textbf{0.747} \scriptsize{(18\%)} & 10.2 & \textbf{0.729} \scriptsize{(31\%)} & 10.0 & \textbf{0.849} \scriptsize{(45\%)} & 10.1 & 0.838 \scriptsize{(55\%)} & 9.0 \\
\bottomrule
\end{tabular*}}%
\begin{threeparttable}
\resizebox{\linewidth}{!}{\usebox\tabmainbox}
\begin{tablenotes}
\scriptsize
\item[$\triangle$]\label{fn:tau} Even a full sweep of $\tau$ for LA-Entropy does not improve over \method   on real-world benchmarks (Figure~\ref{fig:tau_sweep}).
\item[$\ast$]\label{fn:laentropy}LA-Entropy can exceed \textit{Oracle} as it retrains a separate head per layer, while \textit{Oracle} uses the final-layer head like \method.
\end{tablenotes}
\end{threeparttable}
}
\end{table}

\vspace{-.05in}
\subsection{Results}
\vspace{-.075in}

\textbf{Adaptive exits deliver better detection with less computation.}
Table~\ref{tab:main} shows that \method matches or improves Full-depth AUROC in 11 of 12 settings while computing 15--63\% fewer layers, and recovers up to 45\% of the \textit{Oracle} improvement on real-world benchmarks (also see Figure \ref{fig:tradeoff}).

LA-Entropy exits much earlier, at 1.6--2.9 layers, but sacrifices detection performance: despite training a separate prediction head per layer, it underperforms even Half-depth inference for \outformer and \iclad on every benchmark, and Full-depth \tactic on all real-world benchmarks. 

\vspace{-0.25in}
\begin{itemize}[label={},leftmargin=0pt,itemsep=2.5pt,parsep=0pt,partopsep=0pt]
\item %
\item \textit{Beyond Best fixed layer.} \method delivers higher average performance than the \textit{Best fixed layer} (selected using test labels) in 8 of 9 real-world settings,  while reducing its number of computed layers by approximately 15--33\%, highlighting the value of dataset-adaptive exit selection.
\item \textit{Cross-backbone gains on SynBench.} On SynBench, generated from \outformer's priors, \method also captures  early-exit gains for \iclad and \tactic, closing 71\% and 55\% of their Full-depth-to-\textit{Oracle} gaps while computing on average just 4.4 and 9.0 of 12 layers, respectively.
\end{itemize}

\begin{table}[!htbp]
\vspace{-0.05in}
\centering
\caption{Total compute and inference time (s)  for each benchmark on one H100 GPU, full depth vs. \method (router calls and extra features included). %
}
\vspace{-0.05in}
\label{tab:compute}
{\footnotesize\setlength{\tabcolsep}{1.2pt}
\resizebox{\linewidth}{!}{\renewcommand{\arraystretch}{0.942}%
\begin{tabular*}{460pt}{@{}cl@{\extracolsep{\fill}}rrrrrrrrrrrr@{}}
\toprule
 &  & \multicolumn{4}{c}{\small OddBench} & \multicolumn{4}{c}{\small OvRBench} & \multicolumn{4}{c}{\small ADBench} \\
\cmidrule(lr){3-6}\cmidrule(lr){7-10}\cmidrule(lr){11-14}
Backbone & Method & PFLOPs & \begin{tabular}[b]{@{}c@{}}Time\\(s)\end{tabular} & \begin{tabular}[b]{@{}c@{}}\textbf{Speedup}\\($\uparrow$)\end{tabular} & \begin{tabular}[b]{@{}c@{}}AUROC\\\textbf{gain} ($\uparrow$)\end{tabular} & PFLOPs & \begin{tabular}[b]{@{}c@{}}Time\\(s)\end{tabular} & \begin{tabular}[b]{@{}c@{}}\textbf{Speedup}\\($\uparrow$)\end{tabular} & \begin{tabular}[b]{@{}c@{}}AUROC\\\textbf{gain} ($\uparrow$)\end{tabular} & PFLOPs & \begin{tabular}[b]{@{}c@{}}Time\\(s)\end{tabular} & \begin{tabular}[b]{@{}c@{}}\textbf{Speedup}\\($\uparrow$)\end{tabular} & \begin{tabular}[b]{@{}c@{}}AUROC\\\textbf{gain} ($\uparrow$)\end{tabular} \\
\midrule
\multirow{2}{*}{\shortstack{OutFormer\\ ($L{=}10$)}} & Full depth & 2.29 & 111 & -- & -- & 3.07 & 146 & -- & -- & 0.21 & 9.9 & -- & -- \\
 & \textbf{rOUT} & 1.21 & 69.0 & 1.61$\times$ & +2.8\% & 1.48 & 82.4 & 1.77$\times$ & +3.0\% & 0.11 & 6.0 & 1.64$\times$ & +0.9\% \\
\midrule
\multirow{2}{*}{\shortstack{ICLAD\\ ($L{=}12$)}} & Full depth & 2.04 & 70.6 & -- & -- & 2.64 & 89.9 & -- & -- & 0.18 & 6.4 & -- & -- \\
 & \textbf{rOUT} & 1.25 & 53.1 & 1.33$\times$ & +0.8\% & 1.36 & 56.7 & 1.59$\times$ & +1.4\% & 0.087 & 3.8 & 1.67$\times$ & +0.0\% \\
\midrule
\multirow{2}{*}{\shortstack{TACTIC\\ ($L{=}12$)}} & Full depth & 2.71 & 183 & -- & -- & 3.64 & 246 & -- & -- & 0.25 & 17.0 & -- & -- \\
 & \textbf{rOUT} & 2.52 & 178 & 1.03$\times$ & +1.5\% & 3.19 & 226 & 1.09$\times$ & +1.9\% & 0.19 & 13.9 & 1.22$\times$ & +2.1\% \\
\bottomrule
\end{tabular*}
}
}
\vspace{-0.05in}
\end{table}

Table~\ref{tab:compute} shows that
\method accelerates inference across all three backbones and benchmarks, achieving $1.61$--$1.77\times$ speedups for \outformer and $1.33$--$1.67\times$ for \iclad, delivering up to 3.0\% relative AUROC gain.
For \tactic, \method retains more layers when beneficial for detection, achieving up to $1.22\times$ acceleration alongside a 2.1\% relative AUROC gain. These savings account for backbone computation, router calls, and additional-feature computation (breakdown in Table~\ref{tab:compute_detail}).

\begin{wrapfigure}{r}{0.59\linewidth}
\vspace{-0.05in}
\hspace{-0.1in}
\centering
\includegraphics[width=\linewidth]{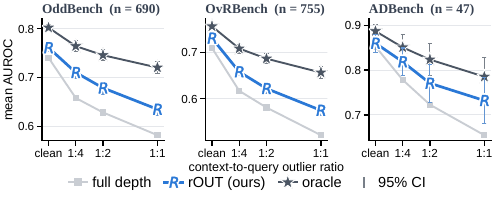}
\vspace{-0.025in}
\caption{\method under \textbf{near-duplicate pollution} improves over Full depth while exiting early  (\outformer).}
\label{fig:pollution_router_sibling} 
\vspace{-0.025in}
\end{wrapfigure}

\vspace{-0.1in}
\textbf{Context pollution amplifies the gains from adaptive exits.}
Table~\ref{tab:pollution_fixedhalf} shows that \method recovers the greater early-exit opportunities under pollution  (Figure~\ref{fig:pollution_router}): on all three real-world benchmarks, its relative AUROC gains over Full-depth \outformer are larger in every tested held-out pollution setting than with clean context (similar results in Table \ref{tab:pollution_fixedhalf_all} for other backbones). On ADBench, the gain rises from 0.9\% with clean context to 5.3\%, 8.6\%, and 10.1\% at context-to-query outlier ratios of 1:4, 1:2, and 1:1, respectively. Gains also reach 8.0\% on OddBench and 6.0\% on OvRBench, compared with 2.8\% and 3.0\% with clean context. \method outperforms both Half depth and LA-Entropy across all tested pollution settings, demonstrating the value of adaptive exit selection when the context itself contains hidden outliers.
Figure \ref{fig:pollution_router_sibling}  mirrors these results under near-duplicate pollution, with even larger gains in this setting. (See Table \ref{tab:pollution_sibling_all} for full results.)

\begin{table}[!htbp]
\centering
\caption{\method under \textbf{held-out pollution} of the context (\outformer).  Columns give the context-to-query outlier ratio; at a 1:1 ratio, the context pollution rate is close to the dataset's natural outlier rate (median 0.8--1.2$\times$).
The row under \method gives its relative AUROC gain over full depth.}
\label{tab:pollution_fixedhalf}
{\footnotesize\setlength{\tabcolsep}{1.6pt}
\begingroup
\small %
\resizebox{\linewidth}{!}{%
\begin{tabular*}{1.1111\linewidth}{@{} l @{\extracolsep{\fill}} ccccccccccc c @{}}
\toprule
{\normalsize\textbf{Held-out Pollution}} & \multicolumn{4}{c}{\normalsize OddBench} & \multicolumn{4}{c}{\normalsize OvRBench} & \multicolumn{4}{c}{\normalsize ADBench} \\
\cmidrule(lr){2-5}\cmidrule(lr){6-9}\cmidrule(lr){10-13}
Method & Clean & 1:4 & 1:2 & 1:1 & Clean & 1:4 & 1:2 & 1:1 & Clean & 1:4 & 1:2 & 1:1 \\
\midrule
\textit{Oracle} & .802 & .757 & .741 & .719 & .756 & .706 & .684 & .657 & .887 & .841 & .813 & .773 \\
\textit{Best fixed layer} & .746 & .666 & .637 & .606 & .713 & .634 & .601 & .563 & .854 & .771 & .726 & .666 \\
\cline{1-13}\noalign{\smallskip}
Full depth & .740 & .646 & .616 & .580 & .709 & .625 & .594 & .556 & .853 & .760 & .704 & .636 \\
Half depth & .745 & .651 & .622 & .587 & .707 & .623 & .594 & .562 & .843 & .747 & .694 & .626 \\
LA-Entropy & .709 & .657 & .639 & .612 & .680 & .627 & .606 & .579 & .770 & .723 & .706 & .689 \\
\textbf{rOUT (ours)} & \textbf{.761} & \textbf{.694} & \textbf{.666} & \textbf{.617} & \textbf{.730} & \textbf{.663} & \textbf{.629} & \textbf{.586} & \textbf{.860} & \textbf{.800} & \textbf{.764} & \textbf{.701} \\
\textit{---\,AUROC gain} & {\fontsize{8}{8}\selectfont +2.8\%} & {\fontsize{8}{8}\selectfont +7.4\%} & {\fontsize{8}{8}\selectfont +8.0\%} & {\fontsize{8}{8}\selectfont +6.5\%} & {\fontsize{8}{8}\selectfont +3.0\%} & {\fontsize{8}{8}\selectfont +6.0\%} & {\fontsize{8}{8}\selectfont +5.9\%} & {\fontsize{8}{8}\selectfont +5.4\%} & {\fontsize{8}{8}\selectfont +0.9\%} & {\fontsize{8}{8}\selectfont +5.3\%} & {\fontsize{8}{8}\selectfont +8.6\%} & {\fontsize{8}{8}\selectfont +10.1\%} \\
\bottomrule
\end{tabular*}}
\endgroup}
\end{table}

\begin{wraptable}{r}{0.50\linewidth}
\vspace{-0.01in}
\centering
\caption{Ablations of \method: relative AUROC change (\%) from the full router at matched depth; Full depth is the backbone without routing; one-sided Wilcoxon $^{*}p{<}.05$, $^{**}p{<}.01$, $^{***}p{<}.001$.}
\label{tab:ablation_split}
{\footnotesize
\setlength{\tabcolsep}{2.5pt}
\ifdefined\abgw\else\newlength\abgw\newlength\ablw\newlength\abtmp\fi
\setlength\ablw{0pt}%
\settowidth\abtmp{\textbf{rOUT (full)}}\ifdim\abtmp>\ablw\setlength\ablw{\abtmp}\fi%
\settowidth\abtmp{w/o additional feat.}\ifdim\abtmp>\ablw\setlength\ablw{\abtmp}\fi%
\settowidth\abtmp{w/o layer attention}\ifdim\abtmp>\ablw\setlength\ablw{\abtmp}\fi%
\ifdefined\abstw\else\newlength\abstw\fi\settowidth\abstw{$^{***}$}%
\begin{tabular*}{\linewidth}{@{}>{\raggedright\arraybackslash}p{\ablw}@{\hspace{5pt}\extracolsep{\fill}}c@{\hspace{\tabcolsep}\hspace{\abstw}\hspace{\tabcolsep}}c@{\hspace{\tabcolsep}\hspace{\abstw}\hspace{\tabcolsep}}c@{\hspace{\abstw}}}
\toprule
\outformer & Odd & OvR & AD \\
\midrule
\textbf{rOUT (full)} & .761 & .730 & .860 \\
\midrule
w/o $\mathcal{L}_{\mathrm{layer}}$ & $-$1.5\rlap{$^{***}$} & $-$2.0\rlap{$^{***}$} & $-$0.5 \\
w/o $\mathcal{L}_{\mathrm{seq}}$ & $-$0.6\rlap{$^{***}$} & $-$0.6\rlap{$^{**}$} & $-$0.4\rlap{$^{*}$} \\
w/o PI & $-$2.0\rlap{$^{***}$} & $-$2.4\rlap{$^{***}$} & $-$1.2\rlap{$^{*}$} \\
w/o layer attention & $-$0.8\rlap{$^{**}$} & $-$0.7\rlap{$^{***}$} & $-$1.2\rlap{$^{*}$} \\
w/o additional feat. & $-$1.7\rlap{$^{***}$} & $-$1.0\rlap{$^{***}$} & $-$1.5\rlap{$^{*}$} \\
\midrule
Full depth & $-$2.7\rlap{$^{***}$} & $-$2.9\rlap{$^{***}$} & $-$0.9\rlap{$^{*}$} \\
\bottomrule
\end{tabular*}
}
\vspace{-0.1in}
\end{wraptable}

\textbf{Privileged information and richer representations strengthen routing.}
Table~\ref{tab:ablation_split} presents component-wise ablations of \method on \outformer. For a fair comparison, each variant's threshold $\tau$ is tuned to match the full router's mean layer budget. The full router  outperforms all ablated variants significantly across all three real-world benchmarks. 
Removing PI causes the largest drop (up to 2.4\% AUROC), followed by omitting either routing loss (up to 2.0\%), additional features (up to 1.7\%), and layer attention (up to 1.2\%).
These consistent performance drops confirm that each component contributes to exit selection. 
 Together, they enable \method to significantly outperform Full depth across all benchmarks by up to 3.0\%.
We further ablate the training data recipes, where different sources benefit different evaluation settings (Table~\ref{tab:ablation_data}). Ablating the depth penalty ($\lambda=0$) costs \outformer 3.3--4.3 more computed layers at nearly unchanged AUROC (Table~\ref{tab:lambda}).

\vspace{-.075in}
\section{Conclusion}
\vspace{-.05in}
\label{sec:conclusion}

We present the first study of early exit for pretrained tabular outlier detection, revealing that it improves not only efficiency but also  performance.
We identify context pollution as a key mechanism: deeper processing strengthens the influence of context outliers that mask query outliers, which early exit mitigates.
Building on this, we introduce \method, which learns sequential stopping and retrospective layer selection using annealed access to privileged query labels during pretraining. Across three backbones and real-world benchmarks, \method achieves up to $1.8\times$ speedup while recovering up to $45\%$ of Oracle gains on clean contexts, with gains growing to  $59\%$ under context pollution.

\bibliography{refs}

\begin{thebibliography}{53}
\providecommand{\natexlab}[1]{#1}
\providecommand{\url}[1]{\texttt{#1}}
\expandafter\ifx\csname urlstyle\endcsname\relax
  \providecommand{\doi}[1]{doi: #1}\else
  \providecommand{\doi}{doi: \begingroup \urlstyle{rm}\Url}\fi

\bibitem[Banino et~al.(2021)Banino, Balaguer, and
  Blundell]{banino2021pondernet}
Andrea Banino, Jan Balaguer, and Charles Blundell.
\newblock {PonderNet}: Learning to ponder.
\newblock In \emph{8th ICML Workshop on Automated Machine Learning}, 2021.
\newblock URL \url{https://arxiv.org/abs/2107.05407}.

\bibitem[Chen et~al.(2026)Chen, Ding, and Akoglu]{chen2026vip}
Yilong Chen, Xueying Ding, and Leman Akoglu.
\newblock {VIP-COP}: Context optimization for tabular foundation models.
\newblock \emph{arXiv preprint arXiv:2605.12904}, 2026.

\bibitem[Dai et~al.(2023)Dai, Sun, Dong, Hao, Ma, Sui, and
  Wei]{dai2023whycangpt}
Damai Dai, Yutao Sun, Li~Dong, Yaru Hao, Shuming Ma, Zhifang Sui, and Furu Wei.
\newblock Why can {GPT} learn in-context? language models secretly perform
  gradient descent as meta-optimizers.
\newblock In \emph{Findings of the Association for Computational Linguistics:
  ACL 2023}, pp.\  4005--4019, 2023.
\newblock \doi{10.18653/v1/2023.findings-acl.247}.

\bibitem[Ding et~al.(2026{\natexlab{a}})Ding, Kl{\"u}ttermann, Wen, Chen, and
  Akoglu]{ding2026macrodata}
Xueying Ding, Simon Kl{\"u}ttermann, Haomin Wen, Yilong Chen, and Leman Akoglu.
\newblock {MacrOData}: New benchmarks of thousands of datasets for tabular
  outlier detection.
\newblock In \emph{Proceedings of the 32nd ACM SIGKDD Conference on Knowledge
  Discovery and Data Mining (KDD '26)}, pp.\  8777--8788, 2026{\natexlab{a}}.
\newblock \doi{10.1145/3770855.3817520}.
\newblock URL \url{https://doi.org/10.1145/3770855.3817520}.

\bibitem[Ding et~al.(2026{\natexlab{b}})Ding, Wen, Kl{\"u}ttermann, and
  Akoglu]{ding2026zero}
Xueying Ding, Haomin Wen, Simon Kl{\"u}ttermann, and Leman Akoglu.
\newblock From zero to hero: Advancing zero-shot foundation models for tabular
  outlier detection.
\newblock In \emph{Proceedings of the 43rd International Conference on Machine
  Learning (ICML)}. PMLR, 2026{\natexlab{b}}.

\bibitem[Du et~al.(2022)Du, Huang, Dai, Tong, Lepikhin, Xu, Krikun, Zhou, Yu,
  Firat, Zoph, Fedus, Bosma, Zhou, Wang, Wang, Webster, Pellat, Robinson,
  Meier-Hellstern, Duke, Dixon, Zhang, Le, Wu, Chen, and Cui]{glamgoogle}
Nan Du, Yanping Huang, Andrew~M Dai, Simon Tong, Dmitry Lepikhin, Yuanzhong Xu,
  Maxim Krikun, Yanqi Zhou, Adams~Wei Yu, Orhan Firat, Barret Zoph, Liam Fedus,
  Maarten Bosma, Zongwei Zhou, Tao Wang, Yu~Emma Wang, Kellie Webster, Marie
  Pellat, Kevin Robinson, Kathleen Meier-Hellstern, Toju Duke, Lucas Dixon, Kun
  Zhang, Quoc~V Le, Yonghui Wu, Zhifeng Chen, and Claire Cui.
\newblock {GL}a{M}: Efficient scaling of language models with
  mixture-of-experts.
\newblock In Kamal Chaudhuri, Stefanie Jegelka, Le~Song, Csaba Szepesvari, Gang
  Niu, and Sivan Sabato (eds.), \emph{Proceedings of the Thirty-Ninth
  International Conference on Machine Learning}, volume 162 of
  \emph{Proceedings of Machine Learning Research}, pp.\  5547--5569. PMLR,
  17--23 Jul 2022.

\bibitem[Elbayad et~al.(2020)Elbayad, Gu, Grave, and Auli]{elbayad2020depth}
Maha Elbayad, Jiatao Gu, Edouard Grave, and Michael Auli.
\newblock Depth-adaptive transformer.
\newblock In \emph{International Conference on Learning Representations}, 2020.

\bibitem[Elhoushi et~al.(2024)Elhoushi, Shrivastava, Liskovich, Hosmer, Wasti,
  Lai, Mahmoud, Acun, Agarwal, Roman, Aly, Chen, and Wu]{elhoushi2024layerskip}
Mostafa Elhoushi, Akshat Shrivastava, Diana Liskovich, Basil Hosmer, Bram
  Wasti, Liangzhen Lai, Anas Mahmoud, Bilge Acun, Saurabh Agarwal, Ahmed Roman,
  Ahmed~A Aly, Beidi Chen, and Carole-Jean Wu.
\newblock {LayerSkip}: Enabling early exit inference and self-speculative
  decoding.
\newblock In \emph{Proceedings of the 62nd Annual Meeting of the Association
  for Computational Linguistics (Volume 1: Long Papers)}, pp.\  12622--12642,
  2024.

\bibitem[Fedus et~al.(2022)Fedus, Zoph, and Shazeer]{Fedus2021SwitchTS}
William Fedus, Barret Zoph, and Noam Shazeer.
\newblock Switch transformers: Scaling to trillion parameter models with simple
  and efficient sparsity.
\newblock \emph{Journal of Machine Learning Research}, 23\penalty0
  (120):\penalty0 1--39, 2022.

\bibitem[Feuer et~al.(2024)Feuer, Schirrmeister, Cherepanova, Hegde, Hutter,
  Goldblum, Cohen, and White]{feuer2024tunetables}
Benjamin Feuer, Robin~Tibor Schirrmeister, Valeriia Cherepanova, Chinmay Hegde,
  Frank Hutter, Micah Goldblum, Niv Cohen, and Colin White.
\newblock {TuneTables}: Context optimization for scalable prior-data fitted
  networks.
\newblock In \emph{Advances in Neural Information Processing Systems},
  volume~37, pp.\  83430--83464, 2024.

\bibitem[Geiping et~al.(2025)Geiping, McLeish, Jain, Kirchenbauer, Singh,
  Bartoldson, Kailkhura, Bhatele, and Goldstein]{geiping2025scaling}
Jonas Geiping, Sean McLeish, Neel Jain, John Kirchenbauer, Siddharth Singh,
  Brian~R. Bartoldson, Bhavya Kailkhura, Abhinav Bhatele, and Tom Goldstein.
\newblock Scaling up test-time compute with latent reasoning: A recurrent depth
  approach.
\newblock In \emph{Advances in Neural Information Processing Systems},
  volume~38, 2025.
\newblock URL \url{https://arxiv.org/abs/2502.05171}.

\bibitem[Giannou et~al.(2023)Giannou, Rajput, Sohn, Lee, Lee, and
  Papailiopoulos]{giannou2023looped}
Angeliki Giannou, Shashank Rajput, Jy-Yong Sohn, Kangwook Lee, Jason~D. Lee,
  and Dimitris Papailiopoulos.
\newblock Looped transformers as programmable computers.
\newblock In \emph{Proceedings of the 40th International Conference on Machine
  Learning}, volume 202 of \emph{Proceedings of Machine Learning Research},
  pp.\  11398--11442. PMLR, 2023.

\bibitem[Grinsztajn et~al.(2026)Grinsztajn, Fl{\"o}ge, Key, Birkel, Jund, Roof,
  Manium, Hoo, B{\"u}hler, Garg, Safaric, Robertson, J{\"a}ger, Alessi, Hayler,
  Moroshan, Purucker, Singer, Arazi, Siems, Metzen, Grab, Erickson, Guo,
  Kalfon, Bing, Salinas, Cornu, Wehrhahn, Kriuchkova, Kaya, Sidhoum, Salmon,
  Chen, Hulsebos, LeCun, M{\"u}ller, Sch{\"o}lkopf, Gambhir, Hollmann, and
  Hutter]{grinsztajn2026tabpfn3}
L{\'e}o Grinsztajn, Klemens Fl{\"o}ge, Oscar Key, Felix Birkel, Philipp Jund,
  Brendan Roof, Mihir Manium, Shi~Bin Hoo, Magnus B{\"u}hler, Anurag Garg,
  Dominik Safaric, Jake Robertson, Benjamin J{\"a}ger, Simone Alessi, Adrian
  Hayler, Vladyslav Moroshan, Lennart Purucker, Philipp Singer, Alan Arazi,
  Julien Siems, Jan~Hendrik Metzen, Georg Grab, Nick Erickson, Siyuan Guo,
  Eliott Kalfon, Simon Bing, David Salinas, Clara Cornu, Lilly~Charlotte
  Wehrhahn, Diana Kriuchkova, Kursat Kaya, Lydia Sidhoum, Marie Salmon, Jerry
  Chen, Madelon Hulsebos, Yann LeCun, Samuel M{\"u}ller, Bernhard
  Sch{\"o}lkopf, Sauraj Gambhir, Noah Hollmann, and Frank Hutter.
\newblock {TabPFN-3}: Technical report.
\newblock \emph{arXiv preprint arXiv:2605.13986}, 2026.
\newblock URL \url{https://arxiv.org/abs/2605.13986}.

\bibitem[Han et~al.(2022)Han, Hu, Huang, Jiang, and Zhao]{han2022adbench}
Songqiao Han, Xiyang Hu, Hailiang Huang, Minqi Jiang, and Yue Zhao.
\newblock {ADBench}: Anomaly detection benchmark.
\newblock In \emph{Advances in Neural Information Processing Systems
  (NeurIPS)}, volume~35, pp.\  32142--32159, 2022.

\bibitem[He et~al.(2025)He, Ge, Sun, Tian, Wang, and Yu]{he2025router}
Shwai He, Tao Ge, Guoheng Sun, Bowei Tian, Xiaoyang Wang, and Dong Yu.
\newblock Router-tuning: A simple and effective approach for dynamic depth.
\newblock In \emph{Proceedings of the 2025 Conference on Empirical Methods in
  Natural Language Processing}, pp.\  1925--1938, 2025.

\bibitem[Heakl et~al.(2026)Heakl, Gubri, Khan, Yun, and Oh]{heakl2026drllm}
Ahmed Heakl, Martin Gubri, Salman Khan, Sangdoo Yun, and Seong~Joon Oh.
\newblock {Dr.LLM}: Dynamic layer routing in {LLMs}.
\newblock In \emph{International Conference on Learning Representations}, 2026.

\bibitem[Hoffmann et~al.(2022)Hoffmann, Borgeaud, Mensch, Buchatskaya, Cai,
  Rutherford, de~Las~Casas, Hendricks, Welbl, Clark,
  et~al.]{hoffmann2022training}
Jordan Hoffmann, Sebastian Borgeaud, Arthur Mensch, Elena Buchatskaya, Trevor
  Cai, Eliza Rutherford, Diego de~Las~Casas, Lisa~Anne Hendricks, Johannes
  Welbl, Aidan Clark, et~al.
\newblock An empirical analysis of compute-optimal large language model
  training.
\newblock In \emph{Advances in Neural Information Processing Systems},
  volume~35, pp.\  30016--30030, 2022.
\newblock \doi{10.52202/068431-2176}.

\bibitem[Hollmann et~al.(2023)Hollmann, M{\"u}ller, Eggensperger, and
  Hutter]{hollmann2023tabpfn}
Noah Hollmann, Samuel M{\"u}ller, Katharina Eggensperger, and Frank Hutter.
\newblock {TabPFN}: A transformer that solves small tabular classification
  problems in a second.
\newblock In \emph{International Conference on Learning Representations
  (ICLR)}, 2023.

\bibitem[Hollmann et~al.(2025)Hollmann, M{\"u}ller, Purucker, Krishnakumar,
  K{\"o}rfer, Hoo, Schirrmeister, and Hutter]{hollmann2025accurate}
Noah Hollmann, Samuel M{\"u}ller, Lennart Purucker, Arjun Krishnakumar, Max
  K{\"o}rfer, Shi~Bin Hoo, Robin~Tibor Schirrmeister, and Frank Hutter.
\newblock Accurate predictions on small data with a tabular foundation model.
\newblock \emph{Nature}, 637\penalty0 (8045):\penalty0 319--326, 2025.

\bibitem[Ilin et~al.(2026)Ilin, Sushko, and Krishna]{ilin2026discoformer}
Vasily Ilin, Peter Sushko, and Ranjay Krishna.
\newblock {DiScoFormer}: Plug-in density and score estimation with
  transformers.
\newblock In \emph{Proceedings of the 43rd International Conference on Machine
  Learning (ICML)}, 2026.
\newblock URL \url{https://arxiv.org/abs/2511.05924}.

\bibitem[Kaplan et~al.(2020)Kaplan, McCandlish, Henighan, Brown, Chess, Child,
  Gray, Radford, Wu, and Amodei]{kaplan2020scaling}
Jared Kaplan, Sam McCandlish, Tom Henighan, Tom~B. Brown, Benjamin Chess, Rewon
  Child, Scott Gray, Alec Radford, Jeffrey Wu, and Dario Amodei.
\newblock Scaling laws for neural language models.
\newblock \emph{arXiv preprint arXiv:2001.08361}, 2020.

\bibitem[Kokhlikyan et~al.(2020)Kokhlikyan, Miglani, Martin, Wang, Alsallakh,
  Reynolds, Melnikov, Kliushkina, Araya, Yan, and
  Reblitz-Richardson]{kokhlikyan2020captum}
Narine Kokhlikyan, Vivek Miglani, Miguel Martin, Edward Wang, Bilal Alsallakh,
  Jonathan Reynolds, Alexander Melnikov, Natalia Kliushkina, Carlos Araya, Siqi
  Yan, and Orion Reblitz-Richardson.
\newblock {Captum}: A unified and generic model interpretability library for
  {PyTorch}, 2020.

\bibitem[Kong \& Das(2026)Kong and Das]{kong2026tabfm}
Weihao Kong and Abhimanyu Das.
\newblock Introducing {TabFM}: A zero-shot foundation model for tabular data.
\newblock Google Research Blog, June 2026.

\bibitem[K{\"u}ken et~al.(2025)K{\"u}ken, Purucker, and Hutter]{kuken2025early}
Jaris K{\"u}ken, Lennart Purucker, and Frank Hutter.
\newblock Early stopping tabular in-context learning.
\newblock In \emph{ICML 2025 Workshop on Foundation Models for Structured
  Data}, 2025.

\bibitem[Lee et~al.(2019)Lee, Lee, Kim, Kosiorek, Choi, and Teh]{lee2019set}
Juho Lee, Yoonho Lee, Jungtaek Kim, Adam Kosiorek, Seungjin Choi, and Yee~Whye
  Teh.
\newblock Set transformer: A framework for attention-based
  permutation-invariant neural networks.
\newblock In \emph{Proceedings of the 36th International Conference on Machine
  Learning}, volume~97 of \emph{Proceedings of Machine Learning Research}, pp.\
   3744--3753. PMLR, 2019.
\newblock URL \url{https://proceedings.mlr.press/v97/lee19d.html}.

\bibitem[Li et~al.(2026)Li, Li, and Zhou]{li2026polar}
Ziyue Li, Yang Li, and Tianyi Zhou.
\newblock Skip a layer or loop it? learning program-of-layers in {LLMs}.
\newblock In \emph{Proceedings of the 43rd International Conference on Machine
  Learning (ICML)}, 2026.
\newblock URL \url{https://arxiv.org/abs/2606.06574}.

\bibitem[Liu et~al.(2020)Liu, Zhou, Wang, Zhao, Deng, and Ju]{liu2020fastbert}
Weijie Liu, Peng Zhou, Zhiruo Wang, Zhe Zhao, Haotang Deng, and Qi~Ju.
\newblock {FastBERT}: a self-distilling {BERT} with adaptive inference time.
\newblock In \emph{Proceedings of the 58th Annual Meeting of the Association
  for Computational Linguistics}, pp.\  6035--6044, 2020.
\newblock \doi{10.18653/v1/2020.acl-main.537}.

\bibitem[Liu et~al.(2021)Liu, Meng, Zhou, Chen, and Xu]{liu2021faster}
Yijin Liu, Fandong Meng, Jie Zhou, Yufeng Chen, and Jinan Xu.
\newblock Faster depth-adaptive transformers.
\newblock In \emph{Proceedings of the AAAI Conference on Artificial
  Intelligence}, pp.\  13424--13432, 2021.

\bibitem[Luo et~al.(2025)Luo, Wang, and Yan]{luo2025adaptive}
Xuan Luo, Weizhi Wang, and Xifeng Yan.
\newblock Adaptive layer-skipping in pre-trained {LLMs}.
\newblock In \emph{Conference on Language Modeling}, 2025.

\bibitem[Ma et~al.(2024)Ma, Thomas, Yu, and Caterini]{ma2024icd}
Junwei Ma, Valentin Thomas, Guangwei Yu, and Anthony Caterini.
\newblock In-context data distillation with {TabPFN}.
\newblock In \emph{ICLR 2024 Workshop on Mathematical and Empirical
  Understanding of Foundation Models (ME-FoMo)}, 2024.
\newblock URL \url{https://arxiv.org/abs/2402.06971}.

\bibitem[Ma et~al.(2025)Ma, Thomas, Hosseinzadeh, Labach, Kamkari, Cresswell,
  Golestan, Yu, Caterini, and Volkovs]{ma2025tabdpt}
Junwei Ma, Valentin Thomas, Rasa Hosseinzadeh, Alex Labach, Hamidreza Kamkari,
  Jesse~C. Cresswell, Keyvan Golestan, Guangwei Yu, Anthony~L. Caterini, and
  Maksims Volkovs.
\newblock {TabDPT}: Scaling tabular foundation models on real data.
\newblock In \emph{Advances in Neural Information Processing Systems}, 2025.
\newblock URL \url{https://arxiv.org/abs/2410.18164}.

\bibitem[Marsza{\l}ek et~al.(2026)Marsza{\l}ek, Ku{\'s}mierczyk, and
  {\'S}mieja]{marszalek2026tactic}
Patryk Marsza{\l}ek, Tomasz Ku{\'s}mierczyk, and Marek {\'S}mieja.
\newblock {TACTIC} for navigating the unknown: Tabular anomaly detection via
  in-context inference.
\newblock \emph{arXiv preprint arXiv:2603.14171}, 2026.

\bibitem[Olsson et~al.(2022)Olsson, Elhage, Nanda, Joseph, DasSarma, Henighan,
  Mann, Askell, Bai, Chen, Conerly, Drain, Ganguli, Hatfield-Dodds, Hernandez,
  Johnston, Jones, Kernion, Lovitt, Ndousse, Amodei, Brown, Clark, Kaplan,
  McCandlish, and Olah]{olsson2022incontext}
Catherine Olsson, Nelson Elhage, Neel Nanda, Nicholas Joseph, Nova DasSarma,
  Tom Henighan, Ben Mann, Amanda Askell, Yuntao Bai, Anna Chen, Tom Conerly,
  Dawn Drain, Deep Ganguli, Zac Hatfield-Dodds, Danny Hernandez, Scott
  Johnston, Andy Jones, Jackson Kernion, Liane Lovitt, Kamal Ndousse, Dario
  Amodei, Tom Brown, Jack Clark, Jared Kaplan, Sam McCandlish, and Chris Olah.
\newblock In-context learning and induction heads.
\newblock \emph{arXiv preprint arXiv:2209.11895}, 2022.
\newblock \doi{10.48550/arxiv.2209.11895}.

\bibitem[Qu et~al.(2025)Qu, Holzm{\"u}ller, Varoquaux, and
  Le~Morvan]{qu2025tabicl}
Jingang Qu, David Holzm{\"u}ller, Ga{\"e}l Varoquaux, and Marine Le~Morvan.
\newblock {TabICL}: A tabular foundation model for in-context learning on large
  data.
\newblock In \emph{International Conference on Machine Learning}, 2025.
\newblock URL \url{https://arxiv.org/abs/2502.05564}.

\bibitem[Qu et~al.(2026)Qu, Holzm{\"u}ller, Varoquaux, and
  Le~Morvan]{qu2026tabiclv2}
Jingang Qu, David Holzm{\"u}ller, Ga{\"e}l Varoquaux, and Marine Le~Morvan.
\newblock {TabICLv2}: A better, faster, scalable, and open tabular foundation
  model.
\newblock In \emph{International Conference on Machine Learning}, 2026.
\newblock URL \url{https://arxiv.org/abs/2602.11139}.

\bibitem[Ramnath et~al.(2025)Ramnath, Zhou, Guan, Mishra, Qi, Shen, Wang, Woo,
  Jeoung, Wang, Wang, Ding, Lu, Xu, Zhou, Srinivasan, Yan, Chen, Ding, Xu, and
  Cheong]{ramnath2025survey}
Kiran Ramnath, Kang Zhou, Sheng Guan, Soumya~Smruti Mishra, Xuan Qi, Zhengyuan
  Shen, Shuai Wang, Sangmin Woo, Sullam Jeoung, Yawei Wang, Haozhu Wang, Han
  Ding, Yuzhe Lu, Zhichao Xu, Yun Zhou, Balasubramaniam Srinivasan, Qiaojing
  Yan, Yueyan Chen, Haibo Ding, Panpan Xu, and Lin~Lee Cheong.
\newblock A systematic survey of automatic prompt optimization techniques.
\newblock In \emph{Proceedings of the 2025 Conference on Empirical Methods in
  Natural Language Processing}, pp.\  33078--33110, 2025.
\newblock \doi{10.18653/v1/2025.emnlp-main.1681}.

\bibitem[Santos et~al.(2026)Santos, Gon{\c{c}}alves, McNamee, Treviso, and
  Martins]{santos2026sparse}
Saul Santos, Nuno Gon{\c{c}}alves, Daniel~C. McNamee, Marcos Treviso, and
  Andr{\'e} F.~T. Martins.
\newblock Sparse attention as compact kernel regression.
\newblock \emph{arXiv preprint arXiv:2601.22766}, 2026.

\bibitem[Schuster et~al.(2022)Schuster, Fisch, Gupta, Dehghani, Bahri, Tran,
  Tay, and Metzler]{schuster2022confident}
Tal Schuster, Adam Fisch, Jai Gupta, Mostafa Dehghani, Dara Bahri, Vinh~Q Tran,
  Yi~Tay, and Donald Metzler.
\newblock Confident adaptive language modeling.
\newblock In \emph{Advances in Neural Information Processing Systems},
  volume~35, pp.\  17456--17472, 2022.

\bibitem[Schwartz et~al.(2020)Schwartz, Stanovsky, Swayamdipta, Dodge, and
  Smith]{schwartz2020right}
Roy Schwartz, Gabriel Stanovsky, Swabha Swayamdipta, Jesse Dodge, and Noah~A.
  Smith.
\newblock The right tool for the job: Matching model and instance complexities.
\newblock In \emph{Proceedings of the 58th Annual Meeting of the Association
  for Computational Linguistics}, pp.\  6640--6651, 2020.
\newblock \doi{10.18653/v1/2020.acl-main.593}.

\bibitem[Shaheen et~al.(2026)Shaheen, Ma, Labach, Hutter, Thomas, and
  Caterini]{shaheen2026understanding}
Nour Shaheen, Junwei Ma, Alex Labach, Frank Hutter, Valentin Thomas, and
  Anthony~L. Caterini.
\newblock Understanding the surprising generalization properties of tabular
  foundation models.
\newblock \emph{arXiv preprint arXiv:2608.17957}, 2026.

\bibitem[Shan et~al.(2024)Shan, Meng, Zheng, Luo, Li, Wang, Xiao, and
  Zhu]{shan2024early}
Weiqiao Shan, Long Meng, Tong Zheng, Yingfeng Luo, Bei Li, Junxin Wang, Tong
  Xiao, and Jingbo Zhu.
\newblock Early exit is a natural capability in transformer-based models: An
  empirical study on early exit without joint optimization, 2024.

\bibitem[Shazeer et~al.(2017)Shazeer, Mirhoseini, Maziarz, Davis, Le, Hinton,
  and Dean]{shazeer2017outrageously}
Noam Shazeer, Azalia Mirhoseini, Krzysztof Maziarz, Andy Davis, Quoc~V Le,
  Geoffrey~E Hinton, and Jeff Dean.
\newblock Outrageously large neural networks: The sparsely-gated
  mixture-of-experts layer.
\newblock In \emph{International Conference on Learning Representations
  (ICLR)}, 2017.

\bibitem[Shen et~al.(2025)Shen, Wen, and Akoglu]{shen2025fomod}
Yuchen Shen, Haomin Wen, and Leman Akoglu.
\newblock {FoMo-0D}: A foundation model for zero-shot tabular outlier
  detection.
\newblock \emph{Transactions on Machine Learning Research}, 2025.
\newblock ISSN 2835-8856.

\bibitem[Skean et~al.(2025)Skean, Arefin, Zhao, Patel, Naghiyev, LeCun, and
  Shwartz-Ziv]{skean2025layer}
Oscar Skean, Md~Rifat Arefin, Dan Zhao, Niket~Nikul Patel, Jalal Naghiyev, Yann
  LeCun, and Ravid Shwartz-Ziv.
\newblock Layer by layer: Uncovering hidden representations in language models.
\newblock In \emph{Proceedings of the 42nd International Conference on Machine
  Learning}, volume 267 of \emph{Proceedings of Machine Learning Research},
  pp.\  55854--55875, 2025.

\bibitem[Snell et~al.(2025)Snell, Lee, Xu, and Kumar]{snell2024scaling}
Charlie Snell, Jaehoon Lee, Kelvin Xu, and Aviral Kumar.
\newblock Scaling {LLM} test-time compute optimally can be more effective than
  scaling parameters for reasoning.
\newblock In \emph{International Conference on Learning Representations
  (ICLR)}, pp.\  10131--10165, 2025.

\bibitem[Sun et~al.(2026)Sun, Yuan, Huang, Zhang, Fu, Wang, Zhou, Li, Wu, and
  Yu]{sun2026survey}
Qingyun Sun, Haonan Yuan, Yi~Huang, Ziwei Zhang, Xingcheng Fu, Ruijie Wang,
  Haoyi Zhou, Jianxin Li, Jia Wu, and Philip~S. Yu.
\newblock A survey on foundation models for structured data: Tabular, time
  series, and graphs.
\newblock \emph{IEEE Transactions on Knowledge and Data Engineering}, 2026.

\bibitem[Vapnik \& Izmailov(2015)Vapnik and Izmailov]{vapnik2015learning}
Vladimir Vapnik and Rauf Izmailov.
\newblock Learning using privileged information: Similarity control and
  knowledge transfer.
\newblock \emph{Journal of Machine Learning Research}, 16\penalty0
  (61):\penalty0 2023--2049, 2015.

\bibitem[Vapnik \& Vashist(2009)Vapnik and Vashist]{vapnik2009new}
Vladimir Vapnik and Akshay Vashist.
\newblock A new learning paradigm: Learning using privileged information.
\newblock \emph{Neural Networks}, 22\penalty0 (5-6):\penalty0 544--557, 2009.

\bibitem[Wei \& Armanfard(2026)Wei and Armanfard]{wei2026iclad}
Jack~Yi Wei and Narges Armanfard.
\newblock {ICLAD}: In-context learning for unified tabular anomaly detection
  across supervision regimes.
\newblock \emph{arXiv preprint arXiv:2603.19497}, 2026.

\bibitem[Xie et~al.(2022)Xie, Raghunathan, Liang, and Ma]{xie2022explanation}
Sang~Michael Xie, Aditi Raghunathan, Percy Liang, and Tengyu Ma.
\newblock An explanation of in-context learning as implicit bayesian inference.
\newblock In \emph{International Conference on Learning Representations
  (ICLR)}, 2022.

\bibitem[Xin et~al.(2020)Xin, Tang, Lee, Yu, and Lin]{xin2020deebert}
Ji~Xin, Raphael Tang, Jaejun Lee, Yaoliang Yu, and Jimmy Lin.
\newblock {DeeBERT}: Dynamic early exiting for accelerating {BERT} inference.
\newblock In \emph{Proceedings of the 58th Annual Meeting of the Association
  for Computational Linguistics}, pp.\  2246--2251, 2020.
\newblock \doi{10.18653/v1/2020.acl-main.204}.

\bibitem[Zhang et~al.(2025)Zhang, Maddix, Yin, Erickson, Ansari, Han, Zhang,
  Akoglu, Faloutsos, Mahoney, Hu, Rangwala, Karypis, and Wang]{zhang2025mitra}
Xiyuan Zhang, Danielle~C. Maddix, Junming Yin, Nick Erickson, Abdul~Fatir
  Ansari, Boran Han, Shuai Zhang, Leman Akoglu, Christos Faloutsos, Michael~W.
  Mahoney, Cuixiong Hu, Huzefa Rangwala, George Karypis, and Bernie Wang.
\newblock {Mitra}: Mixed synthetic priors for enhancing tabular foundation
  models.
\newblock In \emph{Advances in Neural Information Processing Systems}, 2025.

\bibitem[Zhou et~al.(2020)Zhou, Xu, Ge, McAuley, Xu, and Wei]{zhou2020pabee}
Wangchunshu Zhou, Canwen Xu, Tao Ge, Julian McAuley, Ke~Xu, and Furu Wei.
\newblock {BERT} loses patience: Fast and robust inference with early exit.
\newblock In \emph{Advances in Neural Information Processing Systems},
  volume~33, pp.\  18330--18341, 2020.

\end{thebibliography}
\bibliographystyle{iclr2027_conference}

\appendix
\section*{Appendix}

\section{Extended Related Work}
\label{sec:related}

\textbf{Tabular Foundation Models (TFMs).}
Recent advances in tabular machine learning have shifted from dataset-specific training toward amortized in-context inference, establishing Tabular Foundation Models (TFMs) that leverage In-Context Learning (ICL) 
by training Transformers on diverse synthetic priors toward zero-shot inference on unseen tabular datasets in a single forward pass without gradient-based fine-tuning \citep{sun2026survey, hollmann2023tabpfn}. For supervised classification and regression tasks,
building on the pioneering TabPFN~\citep{hollmann2023tabpfn,hollmann2025accurate}, recent advances include scalable inference in TabPFN-3~\citep{grinsztajn2026tabpfn3},  TabICL/TabICLv2~\citep{qu2025tabicl,qu2026tabiclv2}, and TabFM \citep{kong2026tabfm}, diverse synthetic priors in Mitra~\citep{zhang2025mitra}, and pretraining on real-world tables in TabDPT~\citep{ma2025tabdpt}.

Beyond supervised tasks, recent research has successfully adapted the TFM paradigm to unsupervised and zero-shot tabular outlier detection (OD). \textsc{FoMo-0D} introduced the first pretrained outlier detection model by synthetic pretraining with Gaussian mixture priors, enabling zero-shot detection  without additional training or hyperparameter tuning \citep{shen2025fomod}. Its successor \textsc{OutFormer} expanded this direction using diverse synthetic mixtures, including Gaussian mixture, structural causal, and copula-based priors, and introduced a self-evolving curriculum that adaptively allocates training effort across tasks \citep{ding2026zero}. 

More recently, \tactic~\citep{marszalek2026tactic} combines  outlier-centric synthetic priors with contaminated contexts during pretraining to improve robustness to outliers in the context, while \textsc{ICLAD} unifies tabular outlier detection across unsupervised, one-class, and semi-supervised regimes via meta-learned task conditioning \citep{wei2026iclad}.

\textbf{Compute-Adaptive Inference.}
To enable models to dynamically allocate compute, adaptive computation and early-exit mechanisms have emerged. 
 These approaches can be broadly categorized into: (1) designing or training models for adaptive computation via recurrent execution, token-level sparsity, early exit, or layer skipping; and (2) retrofitting existing pretrained backbones through auxiliary modules or post-hoc routing policies.

At the architectural level,
\textbf{natively adaptive designs} include 
Looped Transformers that vary inference-time compute by repeatedly applying shared Transformer blocks, with the number of iterations controlling effective depth~\citep{giannou2023looped,geiping2025scaling}, as well as 
Mixture-of-Experts (MoE) that adapt by employing a gating network to dynamically route each input token to only a small subset of specialized expert sub-networks, effectively decoupling massive parameter capacity from active per-token computational cost \citep{shazeer2017outrageously,Fedus2021SwitchTS,glamgoogle}.
Beyond structural designs, other backbones are often trained or fine-tuned with dynamic compute mechanisms through intermediate supervision and auxiliary prediction heads.
Depth-Adaptive Transformer supervises intermediate predictions and learns halting policies~\citep{elbayad2020depth}, while Faster Depth-Adaptive Transformers estimate token-level depth requirements using mutual information or reconstruction loss~\citep{liu2021faster}.
DeeBERT and FastBERT attach intermediate classifiers and use prediction uncertainty to determine when to exit~\citep{xin2020deebert,liu2020fastbert}, whereas PABEE stops when predictions stabilize across successive layers~\citep{zhou2020pabee}.
CALM combines intermediate-layer supervision with calibrated confidence thresholds for token-level early exiting~\citep{schuster2022confident}.
LayerSkip incorporates layer dropout and early-exit supervision with a shared output head during pretraining, continual pretraining, or fine-tuning~\citep{elhoushi2024layerskip}.

To avoid costly backbone retraining to be adaptive, \textbf{post-hoc retrofitting} methods instead learn lightweight auxiliary modules.
Evidence of early-exit capability without joint backbone optimization~\citep{shan2024early}, together with the transferability of intermediate-layer representations~\citep{skean2025layer}, supports this direction.
Router-Tuning trains lightweight routing modules~\citep{he2025router}, while FlexiDepth additionally trains adapters to compensate for representation changes caused by skipped layers~\citep{luo2025adaptive}.
Dr.LLM and PoLaR extend the execution choices to skipping and repeating pretrained layers, learning routers or execution-program predictors from offline search while keeping the backbone frozen~\citep{heakl2026drllm,li2026polar}.
For TFMs, Early Stopping Tabular In-Context Learning augments a frozen TabPFN backbone with separately trained intermediate decoders and entropy-based stopping~\citep{kuken2025early}.

Our work follows this post-hoc direction, learning a dataset-specific exit router for pretrained tabular outlier detectors while reusing the original prediction head at every exit and keeping all backbone and head parameters fixed.

\textbf{Test-Time Optimization.}
While compute-adaptive inference targets time savings by minimizing compute dynamically based on input difficulty, test-time optimization typically aims to improve prediction performance by expending additional inference compute.

Test-time optimization encompasses strategies that dynamically refine inputs, prompts, or support sets during inference to maximize prediction quality. In Large Language Models (LLMs), test-time compute scaling and search-based prompt optimization allocate extra inference steps or refine prompt structures to elicit higher reasoning capabilities \citep{ramnath2025survey, snell2024scaling}. In Tabular Foundation Models (TFMs), where raw input tables often exceed context limits or contain noisy features, context optimization  methods refine support contexts to align inference inputs with pretraining bounds. Soft prompt tuning techniques like TuneTables and In-Context Distillation optimize compact pseudo-contexts via gradient updates \citep{feuer2024tunetables, ma2024icd}, while hard context selection methods like VIP-COP perform black-box, sample- and feature-level attribution to isolate the most influential predictors \citep{chen2026vip}. 

While traditional test-time optimization trades increased latency for accuracy gains, our work demonstrates that performance enhancements and inference time reduction can be jointly realized in pretrained outlier detection models.
By orchestrating compute-adaptive execution, our approach bypasses the conventional efficiency-performance trade-off.

\section{Technical Details}

\subsection{Planted-Sibling Experiment}
\label{ssec:influence}

\textbf{Setup.} In each real-world benchmark, we take datasets in a fixed random order and keep the first 100 that admit 50 targeted outliers (22 on ADBench). The targets are random query outliers at pairwise distances above $2\delta$, with $\delta$ defined below. Each sibling $\bx_s=\bx_q+0.5\,r_q\mathbf{u}$, where $r_q$ is the distance from $\bx_q$ to its nearest context row and $\mathbf{u}$ is a random unit direction, replaces a random context inlier outside the ten nearest context rows of every target. All three backbones share this setup and the statistics below (Figure~\ref{fig:gradient} for \outformer; Figures~\ref{fig:gradient_iclad} and~\ref{fig:gradient_tactic} for \iclad and \tactic).

\textbf{Influence score.} We compute input gradients with automatic differentiation, following the saliency formulation implemented in Captum~\citep{kokhlikyan2020captum}.
For a targeted query outlier $\bx_q$ and its planted sibling $\bx_s$, we define the influence score at exit layer $k$ as
\begin{equation}
R_k(\bx_q,\bx_s)
= \frac{\delta}{\sigma_k}\,\big\|\nabla_{\bx_s}z_k(\bx_q)\big\|_2,
\label{eq:sibling_influence}
\end{equation}
where $z_k$ is the outlier-class logit minus the inlier-class logit.
The distance scale $\delta$ is the median Euclidean distance from query inliers to their nearest point in the clean context, measured in the preprocessed input space.
The score scale $\sigma_k$ is the standard deviation of $z_k$ over all queries at layer $k$, computed separately for each context condition.
These factors account for differences in input distances across datasets and score scales across layers and conditions.
Under a first-order approximation, $R_k$ is the maximum absolute change in the query score for a sibling perturbation of norm $\delta$, expressed in units of $\sigma_k$.
We take the median over the 50 targeted outliers within each dataset, then the median across datasets.
We apply the same normalization to the other context rows shown in Figure~\ref{fig:gradient}; the nearest-neighbor comparison tracks the same row selected from the clean context before and after sibling planting.

For the detection curves in Figure~\ref{fig:gradient}, we compare the targeted outlier scores before and after sibling planting against a fixed reference: the scores of all query inliers under clean context at the same layer.
This comparison isolates changes in the targeted outlier scores.
We average these AUROC values across datasets; the plotted loss is the clean value minus the value after sibling planting.
The reported Pearson correlation is computed across layers between this mean AUROC loss and the median sibling influence.

\subsection{\method Inputs}
\label{app:features}

\textbf{Backbone representations.} The router reads a random subset of up to 256 context and 1,024 query rows per dataset, which keeps its cost independent of dataset size. Each layer's representation is projected onto 64 principal components fitted on the training data and stored in int8 during training. In pilot experiments, routers trained on these compressed representations performed on par with routers trained on the full-precision ones, and halving the number of query rows read by the router lowered performance only slightly. Besides these representations, the router receives three groups of per-row features.

\textbf{Raw features.} Each row enters with its preprocessed input values, zero-padded to 100 dimensions.

\textbf{Layer-wise outlier scores.} We apply the backbone's scoring head at every layer, rank the resulting scores within each dataset and layer, and map the ranks through the inverse standard normal CDF. Context rows, which the backbone does not score, receive zeros.

\textbf{Geometric features.} We add 29 label-free statistics of each row's neighborhood (Table~\ref{tab:geom_features}). They are measured relative to the context of the same dataset, which makes them comparable across datasets.

\begin{table}[ht]
\centering
\caption{Geometric features of each row. Distances are Euclidean in the preprocessed input space; context rows are scored leave-one-out.}
\label{tab:geom_features}
{\footnotesize\setlength{\tabcolsep}{4pt}
\begin{tabular}{@{}lc>{\raggedright\arraybackslash}p{0.66\linewidth}@{}}
\toprule
Feature & Count & Definition \\
\midrule
Neighbor distances & 21 & For each $k\in\{1,2,5,10,20,50,100\}$: log mean distance to the $k$ nearest context rows, its percentile among context rows, and log mean distance to the $k$ nearest other query rows \\
Distance to center & 2 & Standardized distance from the context mean, as a log value and as a percentile \\
Local density & 4 & For each $k\in\{5,20\}$: local outlier factor and reverse-neighbor count \\
Boundary and duplicate & 2 & Fraction of features at the boundary of the quantile transform; whether an exact copy exists in the context \\
\bottomrule
\end{tabular}}
\end{table}

\subsection{\method Training Objective}
\label{app:objective}

We first complete the routing objective, then define the two privileged-supervision terms, and finally describe the privileged-input schedule. The routing objective adds an entropy bonus, weighted by $\eta$, that keeps the selection distributions from collapsing onto a single layer; with $H$ denoting entropy, the complete routing and privileged-supervision objectives are
\begin{align}
\mathcal{L}_{\mathrm{route}}&=\mathcal{L}_{\mathrm{seq}}+\alpha\mathcal{L}_{\mathrm{layer}}-\frac{\eta}{L}\sum_{k=1}^{L}H(\bp_k),\\
\mathcal{L}_{\mathrm{PI}}&=\beta\mathcal{L}_{\mathrm{row}}+\gamma\mathcal{L}_{\mathrm{sum}}.
\end{align}
The two privileged-supervision terms supervise the router at two granularities. The row-level term $\mathcal{L}_{\mathrm{row}}$ asks the router to predict each query row's outlier label at every depth, so that its representations encode which rows are outliers. The distribution-level term $\mathcal{L}_{\mathrm{sum}}$ asks the rows that the router weights as outliers to match the true outliers as a group, in the mean and spread of their backbone representations at every layer up to the current depth. It thus teaches the router where the outlier population lies in each layer's representation space, the information that exit selection relies on. Removing privileged information altogether costs up to 2.4\% AUROC (Table~\ref{tab:ablation_split}).

Let $q_{k,i}$ be the router's predicted outlier probability for query row $i$ at router depth $k$, $y_i\in\{0,1\}$ its outlier label, and $\mathcal{S}$ the supervised query rows: all rows in the first training stage and the unrevealed rows in the second. The two privileged-supervision terms are
\begin{align}
\mathcal{L}_{\mathrm{row}}&=-\frac{1}{L}\sum_{k=1}^{L}\frac{1}{|\mathcal{S}|}\sum_{i\in\mathcal{S}}\big[y_i\log q_{k,i}+(1-y_i)\log(1-q_{k,i})\big],\\
\mathcal{L}_{\mathrm{sum}}&=\frac{1}{L}\sum_{k=1}^{L}\frac{1}{k}\sum_{l\le k}\frac{1}{2D}\Big(\big\|\tfrac{\hat{\bm\mu}_{k,l}-\bm\mu_l}{\mathbf{s}^{\mu}_l}\big\|^2+\big\|\tfrac{\hat{\bm\sigma}_{k,l}-\bm\sigma_l}{\mathbf{s}^{\sigma}_l}\big\|^2\Big),
\end{align}
where $\hat{\bm\mu}_{k,l}$ and $\hat{\bm\sigma}_{k,l}$ are the mean and standard deviation of the layer-$l$ backbone representations of the query rows, weighted by $q_{k,i}/\sum_{i'}q_{k,i'}$, $\bm\mu_l$ and $\bm\sigma_l$ are the same statistics over the true query outliers, $\mathbf{s}^{\mu}_l,\mathbf{s}^{\sigma}_l$ are fixed per-component scales, $D=64$ is the number of principal components, and division is elementwise. Only layers up to the router depth $k$ enter $\mathcal{L}_{\mathrm{sum}}$. Each term is zero for datasets without supervised rows or a valid outlier summary.
We use $\alpha=1$, $\beta=0.05$, $\gamma=1.5$, and $\lambda=0.0025$ for all backbones, with $\eta=0.02$ for \outformer and \iclad and $\eta=0.06$ for \tactic. The full objective is $\mathcal{L}=\mathcal{L}_{\mathrm{route}}+\mathcal{L}_{\mathrm{PI}}$.

\textbf{Privileged-input schedule.} Training has two stages (Table~\ref{tab:recipe}). The first stage (12 epochs) reveals all query labels and the outlier summary on every dataset. The second stage (8 epochs) warm-starts from the first and withdraws both inputs with cosine schedules: the probability that a dataset receives a privileged input decays from 0.75 to 0 along a half-cosine, and a dataset that receives query labels reveals a fraction of its query rows drawn from $U[0,1]$. The summary is withdrawn over the first 30\% of the second stage and the query labels over the next 40\%, so the last 30\% trains entirely without privileged inputs, as at deployment.

\textbf{Training cost.} On one H100 GPU, training a router takes 4.3 hours for \outformer and 5.0 hours for \iclad and \tactic (Table~\ref{tab:recipe}), far less than the three days that pretraining \outformer takes on four L40S GPUs~\citep{ding2026zero}.

\begin{table}[!ht]
\centering
\caption{Training recipe of \method.}
\label{tab:recipe}
{\footnotesize\setlength{\tabcolsep}{3.0pt}
\begin{tabular}{@{}>{\raggedright\arraybackslash}p{3.8cm}>{\centering\arraybackslash}p{4.5cm}|>{\centering\arraybackslash}p{4.5cm}@{}}
\toprule
& Stage 1 & Stage 2 \\
& \small query labels shown & \small privileged inputs withdrawn \\
\midrule
Epochs & 12 & 8 \\
Batch size (datasets) & 64 & 32 \\
Peak learning rate & $5 \times 10^{-4}$ & $2 \times 10^{-4}$ \\
Warmup steps & 300 & 100 \\
\midrule
\multicolumn{3}{@{}l}{\emph{Training time (hours, one H100)}} \\
\outformer & 2.5 & 1.8 \\
\iclad & 3.0 & 2.0 \\
\tactic & 3.0 & 2.0 \\
\midrule
\multicolumn{3}{@{}l}{\emph{Shared by both stages}} \\
\addlinespace[1pt]
Architecture & \multicolumn{2}{>{\raggedright\arraybackslash}p{9.8cm}@{}}{two blocks of Sample and Layer Attention, $d$ = 256, 4 heads, attention pooling} \\
\addlinespace[2pt]
Input & \multicolumn{2}{>{\raggedright\arraybackslash}p{9.8cm}@{}}{backbone representations and the per-row features of Appendix~\ref{app:features}} \\
\addlinespace[2pt]
Optimizer & \multicolumn{2}{>{\raggedright\arraybackslash}p{9.8cm}@{}}{AdamW, weight decay 0.02, gradient clip 1.0; the rate holds at the peak for 60\% of the steps after warmup, then cosine to 10\% of it} \\
\addlinespace[2pt]
At inference & \multicolumn{2}{>{\raggedright\arraybackslash}p{9.8cm}@{}}{stop at the first layer whose probability passes $\tau$; $\tau$ and the epoch are chosen on synthetic validation data} \\
\bottomrule
\end{tabular}
}
\end{table}

\section{Experiment Details}
\label{app:experiments}

\subsection{Setup}
\label{app:setup}

\textbf{Evaluation benchmarks.} We evaluate on three real-world benchmarks and one synthetic benchmark (Table~\ref{tab:benchmarks}): OddBench and OvRBench with 690 and 755 datasets, the 47 classical tabular datasets of ADBench, and SynBench with 800 datasets generated from the \outformer priors. Our main protocol fills the context with nominal inliers only, which isolates the effect of the pollution we add explicitly. On ADBench, the context holds a random 70\% of the inliers and the query set holds the remaining inliers and all outliers; for AUROC evaluation, datasets are upsampled with replacement to at least 1,000 rows and subsampled to at most 10,000, whereas computation and inference time (Tables~\ref{tab:compute} and~\ref{tab:compute_detail}) are measured on the full datasets. All backbones score each dataset in a single forward pass without context ensembling (the ``w.o.\ Ens'' setting of \citet{ding2026zero}), and \tactic uses its polluted-context checkpoint except in Figure~\ref{fig:gradient_tactic}. For consistency with the synthetic data priors, which generate at most 100 features (Table~\ref{tab:prior_params}), each backbone receives at most 100 features, and \tactic at most 50, its input limit; for datasets with more features, we randomly select features up to this limit. Averages across benchmarks weight each benchmark by its number of datasets.

\textbf{Pollution settings.} Both settings replace context inliers with unlabeled query outliers or their siblings, so the context size is unchanged, and the ratios 1:4, 1:2, and 1:1 give the number of polluting context rows relative to the number of query outliers. In \textit{held-out pollution}, we randomly split the query outliers of each dataset into two halves: one half stays in the query set at every ratio, and the other half is a pool from which 25\%, 50\%, or 100\% is moved into the context. The query inliers are subsampled by the same factor, which keeps the query outlier rate unchanged; the clean columns of Tables~\ref{tab:pollution_fixedhalf} and~\ref{tab:pollution_fixedhalf_all} therefore use the full query set, and the polluted columns the retained half. In \textit{near-duplicate pollution}, the query set is unchanged. We give 25\%, 50\%, or 100\% of the query outliers one sibling each: in the space standardized by the context mean and standard deviation, the outlier is moved in a random direction by $\rho$ times its distance to the nearest context row, with $\rho\sim U(0.1,1)$, and the result is clipped to the range of the context.

\textbf{Router pretraining corpus.} \method is trained on 211,495 synthetic datasets, with 6,053 more for validation and 6,128 held out (Table~\ref{tab:corpus}). They come from two families of priors. The \outformer priors generate outliers through five mechanisms: Gaussian mixtures, two copula variants, and two structural causal model variants. The TabPFN priors produce classification datasets, which we turn into outlier detection tasks by relabeling one or several classes as outliers. We keep such a dataset only if its query outliers are separable from its query inliers, that is, a supervised $k$NN classifier with $k=20$, evaluated leave-one-out on the labeled query rows, reaches an AUROC of at least 0.6. Each family contributes datasets with clean and with polluted context. Table~\ref{tab:prior_params} lists how each dataset is sampled, including its dimensionality, context size, and outlier rate and, for polluted context, the fraction of context rows replaced by outliers and the share of near-duplicates among them.

\begin{table}[!ht]
\centering
\caption{Benchmarks used in this study.}
\label{tab:benchmarks}
{\footnotesize\setlength{\tabcolsep}{4.0pt}
\begin{tabular}{@{}llrccc@{}}
\toprule
& & & \multicolumn{3}{c}{Median per dataset} \\
\cmidrule(lr){4-6}
Benchmark & Tables & Datasets & rows & columns & anomalies \\
\midrule
OddBench & real & 690 & 7,504 & 6 & 9.2\% \\
OvRBench & real & 755 & 5,610 & 10 & 18.7\% \\
ADBench & real & 47 & 4,207 & 19 & 18.2\% \\
SynBench & synthetic & 800 & 3,202 & 49 & 25.8\% \\
\bottomrule
\end{tabular}
}
\end{table}

\begin{table}[!ht]
\centering
\caption{Synthetic training corpus of \method.}
\label{tab:corpus}
{\footnotesize\setlength{\tabcolsep}{4.0pt}
\begin{tabular}{@{}>{\raggedright\arraybackslash}m{2.2cm}>{\raggedright\arraybackslash}m{6.0cm}lrrr@{}}
\toprule
Prior family & Mechanisms & Context & Train & Val & Held out \\
\midrule
OutFormer priors & GMM, Copula-Dependence, Copula-Probability, SCM-Structure, SCM-Measurement & \begin{tabular}[c]{@{}l@{}}clean\\polluted\end{tabular} & \begin{tabular}[c]{@{}r@{}}17,298\\51,878\end{tabular} & \begin{tabular}[c]{@{}r@{}}1,000\\1,000\end{tabular} & \begin{tabular}[c]{@{}r@{}}999\\1,000\end{tabular} \\
\addlinespace[4pt]
TabPFN priors & class relabeling: One-vs-Rest, One-vs-One, Many-vs-Rest & \begin{tabular}[c]{@{}l@{}}clean\\polluted\end{tabular} & \begin{tabular}[c]{@{}r@{}}71,376\\70,943\end{tabular} & \begin{tabular}[c]{@{}r@{}}2,037\\2,016\end{tabular} & \begin{tabular}[c]{@{}r@{}}2,081\\2,048\end{tabular} \\
\cmidrule(l){4-6}
 &  & Total & 211,495 & 6,053 & 6,128 \\
\bottomrule
\end{tabular}
}
\end{table}

\begin{table}[!ht]
\centering
\caption{Sampling parameters of the training corpus.}
\label{tab:prior_params}
\resizebox{\linewidth}{!}{\footnotesize\setlength{\tabcolsep}{5.0pt}%
\begin{tabular}{@{}l l >{\raggedright\arraybackslash}p{6.6cm}@{}}
\toprule
Hyperparameter & Values & Description \\
\midrule
\multicolumn{3}{@{}l}{\emph{Each dataset}} \\
\addlinespace[1pt]
$d$ & $\mathrm{LogUniform}(2, 100)$, rounded & Dimensionality of data \\
$N$ & $5{,}000$ & Rows per dataset, context and query together \\
$n_{\mathrm{ctx}}$ & $U\{500, 750, \ldots, 4{,}750\}$ & Context rows, a uniform draw from the grid with step 250; inliers only in the clean pools \\
$n_{\mathrm{query}}$ & $N - n_{\mathrm{ctx}}$ & Query rows, 250 to 4,500, all of them scored \\
$r$ & $\mathrm{Beta}(1,4)$, capped at $0.5$ & Outlier rate among the query rows \\
\midrule
\multicolumn{3}{@{}l}{\emph{Polluted context, in half the pools}} \\
\addlinespace[1pt]
$\varepsilon$ & $U(0, 0.40)$ & Fraction of context rows replaced by outliers \\
$s$ & $0$ or $U(0.1, 0.5)$, each with probability $\tfrac12$ & Fraction of the replaced rows that are near-duplicates \\
$\rho$ & $U(0.1, 1.0)$ & Near-duplicate offset, in units of the source outlier's distance to its nearest clean context row \\
\bottomrule
\end{tabular}
}
\end{table}

\subsection{Other Experiment Results}
\label{sec:app_figs}

\begin{figure}[!ht]
\centering
\includegraphics[width=\linewidth]{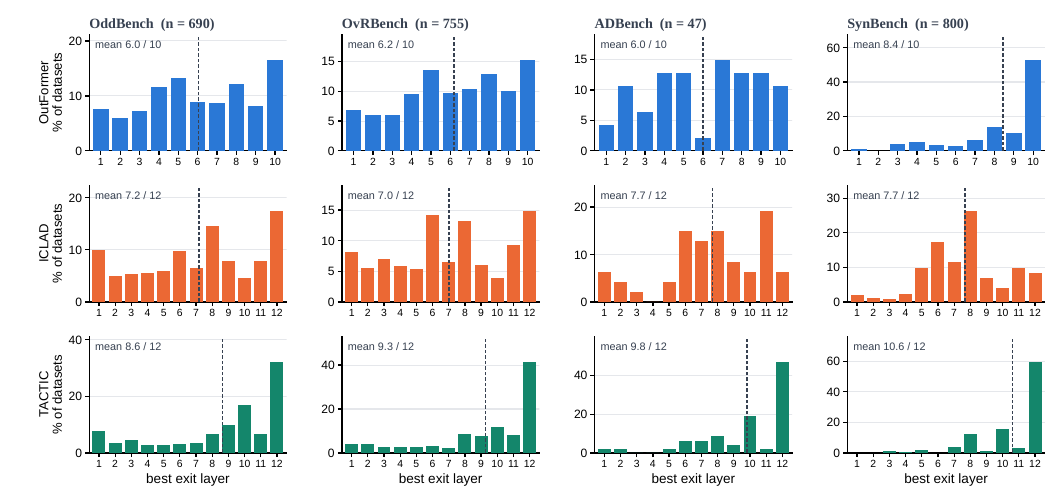}
\caption{Distribution of the oracle exit layer over datasets for every backbone and benchmark; the dashed line marks the mean.}
\label{fig:exit_dist}
\end{figure}

\paragraph{Oracle exit layers.} For every backbone, the oracle exit layer varies widely across datasets. Figure~\ref{fig:exit_dist} shows its per-layer histograms on each benchmark, which underlie the depth groups in Figure~\ref{fig:gain_depth}.

\begin{figure}[!ht]
\centering
\includegraphics[width=\linewidth]{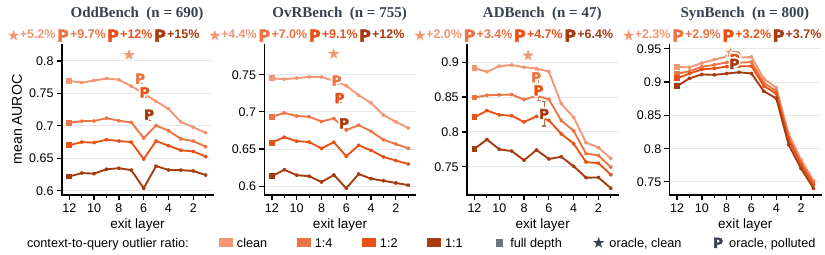}
\caption{Per-layer performance of \iclad under \textbf{held-out pollution}.}
\label{fig:pollution_dose_fixedhalf_iclad}
\end{figure}

\begin{figure}[!ht]
\centering
\includegraphics[width=\linewidth]{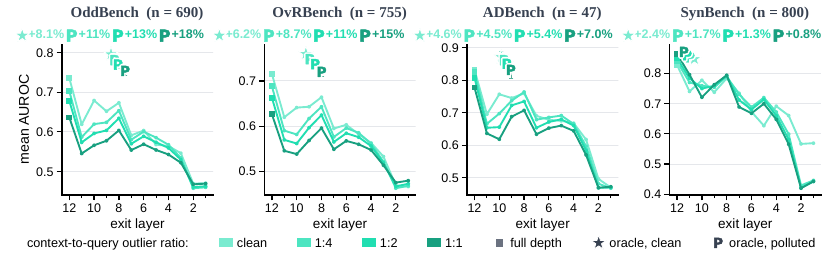}
\caption{Per-layer performance of \tactic under \textbf{held-out pollution}.}
\label{fig:pollution_dose_fixedhalf_tactic}
\end{figure}

\begin{figure}[!ht]
\centering
\includegraphics[width=\linewidth]{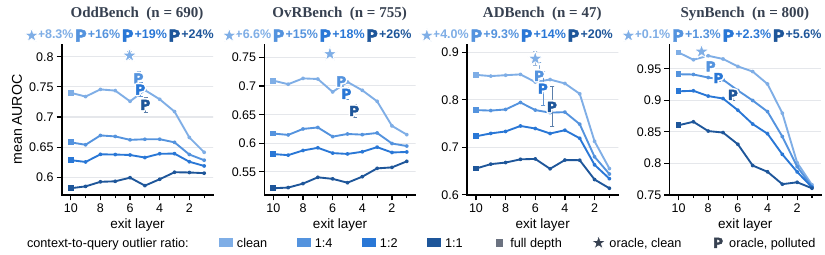}
\caption{Per-layer performance of \outformer under \textbf{near-duplicate pollution}.}
\label{fig:pollution_dose_sibling}
\end{figure}

\begin{figure}[!ht]
\centering
\includegraphics[width=\linewidth]{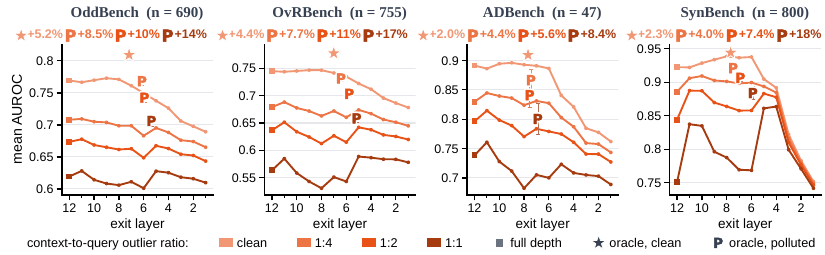}
\caption{Per-layer performance of \iclad under \textbf{near-duplicate pollution}.}
\label{fig:pollution_dose_iclad}
\end{figure}

\begin{figure}[!ht]
\centering
\includegraphics[width=\linewidth]{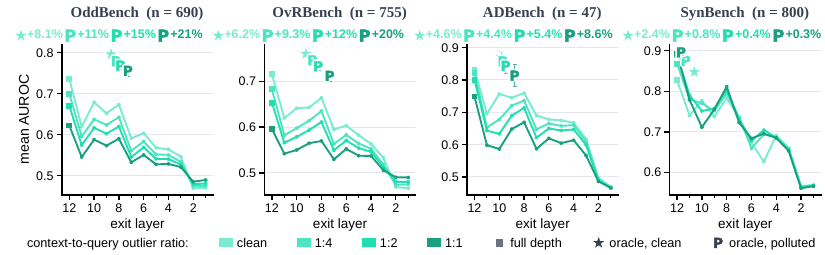}
\caption{Per-layer performance of \tactic under \textbf{near-duplicate pollution}.}
\label{fig:pollution_dose_tactic}
\end{figure}

\FloatBarrier
\paragraph{Context pollution across backbones.} The pollution effect in Figure~\ref{fig:pollution_dose} holds for all three backbones: on the real-world benchmarks, the oracle's gain over full depth grows with the pollution ratio under both pollution modes. Figures~\ref{fig:pollution_dose_fixedhalf_iclad} and~\ref{fig:pollution_dose_fixedhalf_tactic} show held-out pollution for \iclad and \tactic, and Figures~\ref{fig:pollution_dose_sibling}--\ref{fig:pollution_dose_tactic} show near-duplicate pollution for all three backbones.

\begin{figure}[!ht]
\centering
\includegraphics[width=\linewidth]{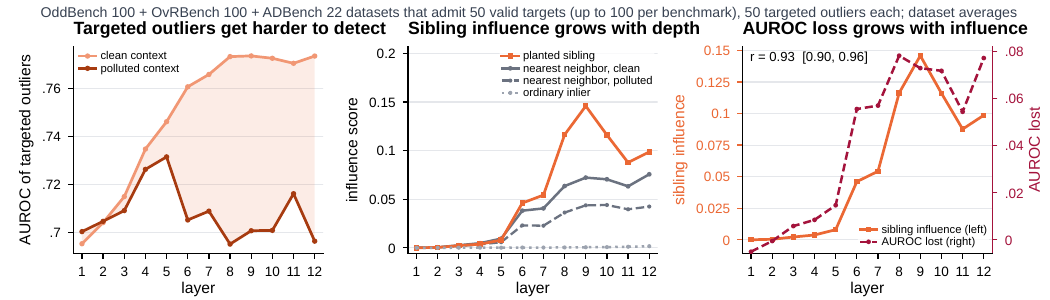}
\caption{One planted sibling per targeted outlier (\iclad).}
\label{fig:gradient_iclad}
\end{figure}

\begin{figure}[!ht]
\centering
\includegraphics[width=\linewidth]{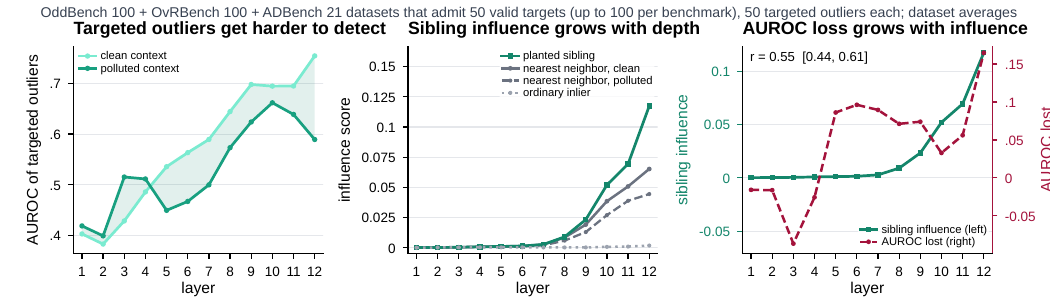}
\caption{One planted sibling per targeted outlier (\tactic, clean-context checkpoint).}
\label{fig:gradient_tactic}
\end{figure}

\FloatBarrier
\paragraph{Planted siblings across backbones.} \iclad and \tactic follow the same mechanism as \outformer, only at different depths (Figures~\ref{fig:gradient_iclad} and~\ref{fig:gradient_tactic}). Sibling influence begins to rise after layer 5 in \iclad but only in the last few layers of \tactic, and in both backbones it tracks the AUROC loss of the targeted outliers across layers (Pearson $r=0.93$ and $0.55$, respectively).

\begin{figure}[!ht]
\centering
\includegraphics[width=\linewidth]{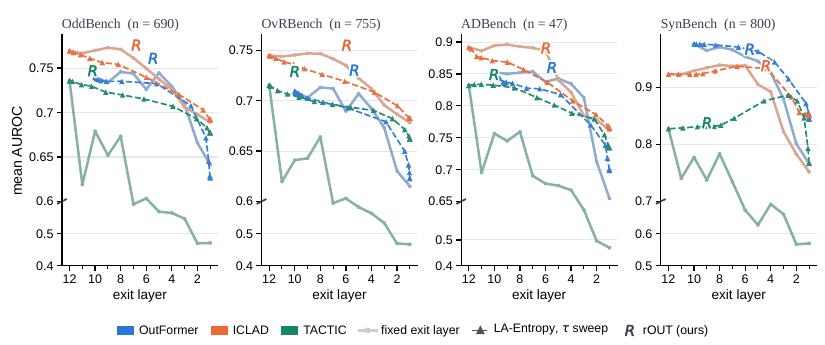}
\caption{LA-Entropy over its full threshold sweep, with exit heads averaged over three seeds, compared with fixed exit layers and \method; its layer-specific heads lift \tactic's early layers well above its fixed-layer curve.}
\label{fig:tau_sweep}
\end{figure}

\begin{figure}[!ht]
\vspace{0.1in}
\centering
\includegraphics[width=\linewidth]{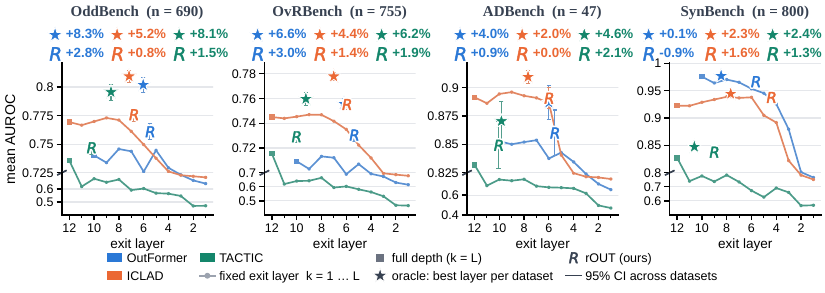}
\caption{Mean AUROC of fixed exit layers, Full depth, the \textit{Oracle}, and \method for every backbone and benchmark (Table~\ref{tab:main}).}
\label{fig:tradeoff}
\vspace{0.2in}
\end{figure}

\FloatBarrier
\paragraph{Baseline and main results.} On the real-world benchmarks, \method lies above the full LA-Entropy threshold sweep for every backbone (Figure~\ref{fig:tau_sweep}). Figure~\ref{fig:tradeoff} shows where \method sits between Full depth and the \textit{Oracle} on each benchmark. Table~\ref{tab:compute_detail} breaks the totals of Table~\ref{tab:compute} down into backbone computation, router calls, and additional features; the latter two account for at most 6.1\% of \method's PFLOPs, so nearly all of the savings come from skipping backbone layers.

\begin{table}[!ht]
\centering
\caption{Full cost breakdown behind Table~\ref{tab:compute}: totals over each benchmark on one H100.}
\label{tab:compute_detail}
{\footnotesize\setlength{\tabcolsep}{4.0pt}
\begin{tabular}{@{}clrrrrrr@{}}
\toprule
 &  & \multicolumn{2}{c}{OddBench} & \multicolumn{2}{c}{OvRBench} & \multicolumn{2}{c}{ADBench} \\
\cmidrule(lr){3-4}\cmidrule(lr){5-6}\cmidrule(lr){7-8}
Backbone &  & PFLOPs & Time (s) & PFLOPs & Time (s) & PFLOPs & Time (s) \\
\midrule
\multirow{6}{*}{\shortstack{OutFormer\\ ($L{=}10$)}} & Full depth & 2.29 & 111 & 3.07 & 146 & 0.21 & 9.9 \\
\noalign{\smallskip}
 & rOUT total & 1.21 & 69.0 & 1.48 & 82.4 & 0.11 & 6.0 \\
 & \quad backbone to exit & 1.15 & 57.8 & 1.42 & 70.5 & 0.11 & 5.1 \\
 & \quad router calls & 0.051 & 7.6 & 0.055 & 8.1 & 0.003 & 0.5 \\
 & \quad extra features & 0.006 & 3.6 & 0.006 & 3.7 & 0.000 & 0.4 \\
\cline{2-8}\noalign{\smallskip}
 & \textbf{Saving} & 47.2\% & 37.9\% & 51.7\% & 43.6\% & 47.9\% & 39.1\% \\
\midrule
\multirow{6}{*}{\shortstack{ICLAD\\ ($L{=}12$)}} & Full depth & 2.04 & 70.6 & 2.64 & 89.9 & 0.18 & 6.4 \\
\noalign{\smallskip}
 & rOUT total & 1.25 & 53.1 & 1.36 & 56.7 & 0.087 & 3.8 \\
 & \quad backbone to exit & 1.17 & 41.2 & 1.29 & 45.3 & 0.083 & 3.0 \\
 & \quad router calls & 0.070 & 8.1 & 0.065 & 7.7 & 0.003 & 0.5 \\
 & \quad extra features & 0.006 & 3.7 & 0.006 & 3.8 & 0.000 & 0.3 \\
\cline{2-8}\noalign{\smallskip}
 & \textbf{Saving} & 38.9\% & 24.8\% & 48.3\% & 36.9\% & 51.1\% & 40.2\% \\
\midrule
\multirow{6}{*}{\shortstack{TACTIC\\ ($L{=}12$)}} & Full depth & 2.71 & 183 & 3.64 & 246 & 0.25 & 17.0 \\
\noalign{\smallskip}
 & rOUT total & 2.52 & 178 & 3.19 & 226 & 0.19 & 13.9 \\
 & \quad backbone to exit & 2.41 & 163 & 3.06 & 208 & 0.18 & 12.6 \\
 & \quad router calls & 0.11 & 12.3 & 0.12 & 14.7 & 0.006 & 0.9 \\
 & \quad extra features & 0.006 & 3.5 & 0.006 & 3.9 & 0.000 & 0.4 \\
\cline{2-8}\noalign{\smallskip}
 & \textbf{Saving} & 6.7\% & 2.5\% & 12.4\% & 8.0\% & 23.9\% & 18.2\% \\
\bottomrule
\end{tabular}
}
\end{table}

\begin{figure}[H]
\centering
\includegraphics[width=0.8\linewidth]{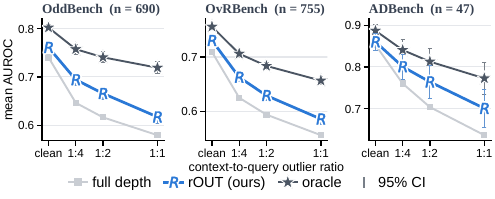}
\caption{\method under \textbf{held-out pollution} improves over Full depth while exiting early (\outformer).}
\label{fig:pollution_router}
\end{figure}

\FloatBarrier
\paragraph{\method under context pollution.} Under both pollution modes, \method gains the most over full depth on \outformer, less on \iclad, and the least on \tactic, which already exits close to full depth. Tables~\ref{tab:pollution_fixedhalf_all} and~\ref{tab:pollution_sibling_all} give the full results for all three backbones, and Figure~\ref{fig:pollution_router} is the held-out counterpart of Figure~\ref{fig:pollution_router_sibling}.

\begin{table}[!ht]
\centering
\caption{\textbf{Held-out pollution} on all three backbones. Columns give the context-to-query outlier ratio; the row under \method gives its relative AUROC gain over full depth.}
\label{tab:pollution_fixedhalf_all}
{\footnotesize\setlength{\tabcolsep}{1.6pt}
\resizebox{\linewidth}{!}{\begin{tabular}{@{}clcccccccccccc@{}}
\toprule
\multicolumn{2}{@{}l}{\textbf{Held-out Pollution}} & \multicolumn{4}{c}{OddBench} & \multicolumn{4}{c}{OvRBench} & \multicolumn{4}{c}{ADBench} \\
\cmidrule(lr){3-6}\cmidrule(lr){7-10}\cmidrule(lr){11-14}
Backbone & Method & Clean & 1:4 & 1:2 & 1:1 & Clean & 1:4 & 1:2 & 1:1 & Clean & 1:4 & 1:2 & 1:1 \\
\midrule
\multirow{7}{*}{\shortstack{OutFormer\\ ($L{=}10$)}} & \textit{Oracle} & .802 & .757 & .741 & .719 & .756 & .706 & .684 & .657 & .887 & .841 & .813 & .773 \\
 & \textit{Best fixed layer} & .746 & .666 & .637 & .606 & .713 & .634 & .601 & .563 & .854 & .771 & .726 & .666 \\
\cline{2-14}\noalign{\smallskip}
 & Full depth & .740 & .646 & .616 & .580 & .709 & .625 & .594 & .556 & .853 & .760 & .704 & .636 \\
 & Half depth & .745 & .651 & .622 & .587 & .707 & .623 & .594 & .562 & .843 & .747 & .694 & .626 \\
 & LA-Entropy & .709 & .657 & .639 & .612 & .680 & .627 & .606 & .579 & .770 & .723 & .706 & .689 \\
 & \textbf{rOUT (ours)} & \textbf{.761} & \textbf{.694} & \textbf{.666} & \textbf{.617} & \textbf{.730} & \textbf{.663} & \textbf{.629} & \textbf{.586} & \textbf{.860} & \textbf{.800} & \textbf{.764} & \textbf{.701} \\
 & \textit{---\,AUROC gain} & {\scriptsize +2.8\%} & {\scriptsize +7.4\%} & {\scriptsize +8.0\%} & {\scriptsize +6.5\%} & {\scriptsize +3.0\%} & {\scriptsize +6.0\%} & {\scriptsize +5.9\%} & {\scriptsize +5.4\%} & {\scriptsize +0.9\%} & {\scriptsize +5.3\%} & {\scriptsize +8.6\%} & {\scriptsize +10.1\%} \\
\midrule
\multirow{7}{*}{\shortstack{ICLAD\\ ($L{=}12$)}} & \textit{Oracle} & .809 & .773 & .752 & .717 & .778 & .742 & .718 & .685 & .909 & .878 & .860 & .826 \\
 & \textit{Best fixed layer} & .773 & .712 & .679 & .638 & .747 & .699 & .666 & .622 & .896 & .854 & .831 & .789 \\
\cline{2-14}\noalign{\smallskip}
 & Full depth & .769 & .705 & .670 & .622 & .745 & .693 & .659 & .613 & .891 & \textbf{.849} & .821 & .776 \\
 & Half depth & .750 & .681 & .649 & .604 & .735 & .675 & .640 & .597 & .887 & .847 & .817 & .761 \\
 & LA-Entropy & .704 & .676 & .659 & .629 & .696 & .662 & .639 & .607 & .786 & .768 & .753 & .732 \\
 & \textbf{rOUT (ours)} & \textbf{.776} & \textbf{.724} & \textbf{.693} & \textbf{.640} & \textbf{.755} & \textbf{.706} & \textbf{.671} & \textbf{.624} & \textbf{.891} & .846 & \textbf{.831} & \textbf{.785} \\
 & \textit{---\,AUROC gain} & {\scriptsize +0.8\%} & {\scriptsize +2.7\%} & {\scriptsize +3.4\%} & {\scriptsize +2.9\%} & {\scriptsize +1.4\%} & {\scriptsize +1.8\%} & {\scriptsize +1.9\%} & {\scriptsize +1.7\%} & {\scriptsize +0.0\%} & {\scriptsize $-$0.4\%} & {\scriptsize +1.2\%} & {\scriptsize +1.2\%} \\
\midrule
\multirow{7}{*}{\shortstack{TACTIC\\ ($L{=}12$)}} & \textit{Oracle} & .796 & .781 & .769 & .754 & .760 & .748 & .737 & .720 & .870 & .862 & .850 & .831 \\
 & \textit{Best fixed layer} & .736 & .703 & .678 & .636 & .715 & .688 & .663 & .626 & .832 & .825 & .807 & .777 \\
\cline{2-14}\noalign{\smallskip}
 & Full depth & .736 & .703 & .678 & \textbf{.636} & .715 & .688 & .663 & \textbf{.626} & .832 & \textbf{.825} & \textbf{.807} & \textbf{.777} \\
 & Half depth & .603 & .601 & .589 & .569 & .603 & .596 & .584 & .567 & .678 & .685 & .671 & .653 \\
 & LA-Entropy & .693 & .657 & .641 & .615 & .682 & .646 & .627 & .599 & .780 & .760 & .747 & .727 \\
 & \textbf{rOUT (ours)} & \textbf{.747} & \textbf{.712} & \textbf{.686} & .634 & \textbf{.729} & \textbf{.698} & \textbf{.670} & .622 & \textbf{.849} & .813 & .802 & .767 \\
 & \textit{---\,AUROC gain} & {\scriptsize +1.5\%} & {\scriptsize +1.2\%} & {\scriptsize +1.2\%} & {\scriptsize $-$0.3\%} & {\scriptsize +1.9\%} & {\scriptsize +1.4\%} & {\scriptsize +1.1\%} & {\scriptsize $-$0.6\%} & {\scriptsize +2.1\%} & {\scriptsize $-$1.4\%} & {\scriptsize $-$0.6\%} & {\scriptsize $-$1.3\%} \\
\bottomrule
\end{tabular}
}}
\end{table}

\begin{table}[!ht]
\centering
\caption{\textbf{Near-duplicate pollution} on all three backbones. Columns give the context-to-query outlier ratio; the row under \method gives its relative AUROC gain over full depth.}
\label{tab:pollution_sibling_all}
{\footnotesize\setlength{\tabcolsep}{1.6pt}
\resizebox{\linewidth}{!}{\begin{tabular}{@{}clcccccccccccc@{}}
\toprule
\multicolumn{2}{@{}l}{\textbf{Near-duplicate Pollution}} & \multicolumn{4}{c}{OddBench} & \multicolumn{4}{c}{OvRBench} & \multicolumn{4}{c}{ADBench} \\
\cmidrule(lr){3-6}\cmidrule(lr){7-10}\cmidrule(lr){11-14}
Backbone & Method & Clean & 1:4 & 1:2 & 1:1 & Clean & 1:4 & 1:2 & 1:1 & Clean & 1:4 & 1:2 & 1:1 \\
\midrule
\multirow{7}{*}{\shortstack{OutFormer\\ ($L{=}10$)}} & \textit{Oracle} & .802 & .764 & .745 & .720 & .756 & .708 & .686 & .656 & .887 & .851 & .823 & .785 \\
 & \textit{Best fixed layer} & .746 & .669 & .639 & .608 & .713 & .628 & .593 & .568 & .854 & .794 & .745 & .675 \\
\cline{2-14}\noalign{\smallskip}
 & Full depth & .740 & .658 & .628 & .582 & .709 & .617 & .581 & .521 & .853 & .778 & .723 & .655 \\
 & Half depth & .745 & .663 & .633 & .586 & .707 & .616 & .581 & .531 & .843 & .773 & .729 & .654 \\
 & LA-Entropy & .709 & .658 & .635 & .600 & .680 & .623 & .601 & .565 & .770 & .737 & .717 & .688 \\
 & \textbf{rOUT (ours)} & \textbf{.761} & \textbf{.710} & \textbf{.678} & \textbf{.635} & \textbf{.730} & \textbf{.658} & \textbf{.622} & \textbf{.575} & \textbf{.860} & \textbf{.819} & \textbf{.771} & \textbf{.732} \\
 & \textit{---\,AUROC gain} & {\scriptsize +2.8\%} & {\scriptsize +7.9\%} & {\scriptsize +8.0\%} & {\scriptsize +9.1\%} & {\scriptsize +3.0\%} & {\scriptsize +6.6\%} & {\scriptsize +7.0\%} & {\scriptsize +10.3\%} & {\scriptsize +0.9\%} & {\scriptsize +5.2\%} & {\scriptsize +6.6\%} & {\scriptsize +11.8\%} \\
\midrule
\multirow{7}{*}{\shortstack{ICLAD\\ ($L{=}12$)}} & \textit{Oracle} & .809 & .768 & .742 & .706 & .778 & .732 & .704 & .659 & .909 & .867 & .841 & .800 \\
 & \textit{Best fixed layer} & .773 & .709 & .678 & .628 & .747 & .688 & .651 & .588 & .896 & .845 & .815 & .761 \\
\cline{2-14}\noalign{\smallskip}
 & Full depth & .769 & .708 & .673 & .619 & .745 & .679 & .637 & .563 & .891 & .830 & .796 & .738 \\
 & Half depth & .750 & .683 & .649 & .601 & .735 & .660 & .615 & .543 & .887 & .827 & .779 & .700 \\
 & LA-Entropy & .704 & .673 & .651 & .615 & .696 & .656 & .629 & .591 & .786 & .761 & .742 & .719 \\
 & \textbf{rOUT (ours)} & \textbf{.776} & \textbf{.724} & \textbf{.693} & \textbf{.637} & \textbf{.755} & \textbf{.693} & \textbf{.655} & \textbf{.594} & \textbf{.891} & \textbf{.844} & \textbf{.806} & \textbf{.756} \\
 & \textit{---\,AUROC gain} & {\scriptsize +0.8\%} & {\scriptsize +2.3\%} & {\scriptsize +2.9\%} & {\scriptsize +2.9\%} & {\scriptsize +1.4\%} & {\scriptsize +2.0\%} & {\scriptsize +2.9\%} & {\scriptsize +5.4\%} & {\scriptsize +0.0\%} & {\scriptsize +1.8\%} & {\scriptsize +1.3\%} & {\scriptsize +2.3\%} \\
\midrule
\multirow{7}{*}{\shortstack{TACTIC\\ ($L{=}12$)}} & \textit{Oracle} & .796 & .779 & .768 & .756 & .760 & .746 & .733 & .712 & .870 & .857 & .843 & .813 \\
 & \textit{Best fixed layer} & .736 & .700 & .670 & .622 & .715 & .682 & .653 & .595 & .832 & .821 & .800 & .749 \\
\cline{2-14}\noalign{\smallskip}
 & Full depth & .736 & .700 & .670 & .622 & .715 & .682 & .653 & .595 & .832 & .821 & .800 & .749 \\
 & Half depth & .603 & .583 & .569 & .551 & .603 & .583 & .571 & .552 & .678 & .665 & .650 & .619 \\
 & LA-Entropy & .693 & .670 & .651 & .619 & .682 & .656 & .636 & .599 & .780 & .766 & .757 & .724 \\
 & \textbf{rOUT (ours)} & \textbf{.747} & \textbf{.716} & \textbf{.686} & \textbf{.645} & \textbf{.729} & \textbf{.692} & \textbf{.667} & \textbf{.616} & \textbf{.849} & \textbf{.822} & \textbf{.811} & \textbf{.759} \\
 & \textit{---\,AUROC gain} & {\scriptsize +1.5\%} & {\scriptsize +2.3\%} & {\scriptsize +2.4\%} & {\scriptsize +3.7\%} & {\scriptsize +1.9\%} & {\scriptsize +1.5\%} & {\scriptsize +2.0\%} & {\scriptsize +3.4\%} & {\scriptsize +2.1\%} & {\scriptsize +0.2\%} & {\scriptsize +1.3\%} & {\scriptsize +1.4\%} \\
\bottomrule
\end{tabular}
}}
\end{table}

\FloatBarrier
\subsection{Additional Ablations}
\label{app:data_ablation}

Table~\ref{tab:ablation_data} extends the component ablations of Section~\ref{sec:exp} to the training data. Each variant drops one data source while keeping the total number of training datasets fixed, and is evaluated at the full router's mean computation depth. Each source matters most in the setting it resembles: without polluted contexts, AUROC drops by 4.4\% on polluted OvRBench; without the TabPFN priors, whose relabeled classes mirror how OvRBench defines outliers, it drops by 2.6\% on clean OvRBench; and without the \outformer priors, it drops by 2.3\% on SynBench. Since no removal improves any other setting by more than 0.5\%, we train on all sources.

\begin{table}[!ht]
\centering
\caption{\textbf{Training-data ablations} of \method on \outformer: relative AUROC change from the full router at matched depth. Poll.\ OvR averages both pollution modes at 1:1; shading marks each source's largest loss; stars as in Table~\ref{tab:ablation_split}.}
\label{tab:ablation_data}
{\footnotesize
\setlength{\tabcolsep}{6pt}
\begin{tabular}{@{}llccc@{}}
\toprule
 &  &  & Poll. &  \\
\multicolumn{2}{@{}l}{Backbone: OutFormer} & OvR & OvR & Syn \\
\midrule
 & \textbf{rOUT (full)} & .730 & .580 & .967 \\
\midrule
\multirow{3}{*}{Data recipe} & w/o polluted ctx. & +0.4\% & {\setlength{\fboxsep}{1.2pt}\colorbox{black!10}{$-$4.4\%\rlap{$^{***}$}}} & 0.0\% \\
 & w/o TabPFN priors & {\setlength{\fboxsep}{1.2pt}\colorbox{black!10}{$-$2.6\%\rlap{$^{***}$}}} & $-$0.5\% & +0.1\% \\
 & w/o OutFormer priors & 0.0\% & +0.5\% & {\setlength{\fboxsep}{1.2pt}\colorbox{black!10}{$-$2.3\%\rlap{$^{***}$}}} \\
\bottomrule
\end{tabular}
}
\vspace{0.2in}
\end{table}

Table~\ref{tab:lambda} ablates the depth penalty. Because the regret $R_j$ is measured in AUROC, $\lambda$ acts as the price of one more layer: at $\lambda=0.0025$, the router goes deeper only when it expects the AUROC gain to outweigh this price. Without the penalty, extra layers cost nothing and an earlier layer can still be selected retrospectively, so the router runs close to full depth, computing 3.3--4.3 more layers for \outformer and up to 7.3 more for \iclad, while gaining at most 0.006 AUROC.

\begin{table}[!ht]
\centering
\caption{\textbf{Depth penalty ablations}: \method with depth cost $\lambda=0.0025$ (default) vs.\ $\lambda=0$: AUROC and mean exit layer.}
\label{tab:lambda}
{\footnotesize\setlength{\tabcolsep}{4.0pt}
\begin{tabular}{@{}llcccccccc@{}}
\toprule
 & & \multicolumn{2}{c}{OddBench} & \multicolumn{2}{c}{OvRBench} & \multicolumn{2}{c}{ADBench} & \multicolumn{2}{c}{SynBench} \\
\cmidrule(lr){3-4}\cmidrule(lr){5-6}\cmidrule(lr){7-8}\cmidrule(lr){9-10}
Backbone & $\lambda$ & AUROC & Layers & AUROC & Layers & AUROC & Layers & AUROC & Layers \\
\midrule
\multirow{2}{*}{\shortstack[l]{OutFormer\\($L$=10)}} & 0.0025 & .761 & 5.5 & .730 & 5.4 & .860 & 5.5 & .967 & 5.7 \\
 & 0 & .761 & 8.8 & .731 & 8.7 & .856 & 9.2 & .973 & 10.0 \\
\midrule
\multirow{2}{*}{\shortstack[l]{ICLAD\\($L$=12)}} & 0.0025 & .776 & 6.8 & .755 & 6.0 & .891 & 6.0 & .938 & 4.4 \\
 & 0 & .777 & 11.1 & .757 & 11.0 & .896 & 11.3 & .939 & 11.7 \\
\midrule
\multirow{2}{*}{\shortstack[l]{TACTIC\\($L$=12)}} & 0.0025 & .747 & 10.2 & .729 & 10.0 & .849 & 10.1 & .838 & 9.0 \\
 & 0 & .745 & 11.6 & .732 & 11.6 & .852 & 11.4 & .840 & 11.6 \\
\bottomrule
\end{tabular}
}
\end{table}

\end{document}